\documentclass[preprint]{uai2026}

\usepackage[T1]{fontenc}                       
\usepackage[american]{babel}                   
\usepackage{natbib}                            
\usepackage{url}                               

\usepackage{amsthm,amsmath,amssymb,amsfonts}   
\usepackage{thmtools}                          
\usepackage{thm-restate}                       

\usepackage{algorithm,algorithmic}             
\usepackage{tabularx}                          
\usepackage{longtable}                         
\usepackage{sidecap}                           
\usepackage{subcaption}                        

\usepackage{bm}                                
\usepackage{xspace}                            

\usepackage{xcolor}                            
\usepackage[many]{tcolorbox}                   
\usepackage{framed}                            

\usepackage{tikz}                              
\usetikzlibrary{arrows.meta,positioning,calc,shapes.geometric,shapes.misc,fit}

\usepackage{hyperref}                          

\newcommand{\I}{\bm{I}}
\DeclareRobustCommand{\wh}[1]{\ensuremath{\widehat{#1}}}
\DeclareRobustCommand{\wt}[1]{\ensuremath{\widetilde{#1}}}

\DeclareMathOperator{\tr}{Tr}
\DeclareMathOperator{\var}{Var}
\DeclareMathOperator{\cov}{Cov}
\DeclareMathOperator{\diag}{diag}
\DeclareMathOperator{\sign}{sign}

\newcommand{\eg}{e.g.\xspace}
\newcommand{\ie}{i.e.\xspace}
\newcommand{\iid}{i.i.d.\xspace}
\newcommand{\vs}{vs.\xspace}
\newcommand{\resp}{resp.\xspace}
\newcommand{\wrt}{w.r.t.\xspace}

\definecolor{linkcolor}{RGB}{44, 96, 163}

\definecolor{color_all_target}{HTML}{4E79A7}
\definecolor{color_all_target_source}{HTML}{B07AA1}
\definecolor{color_oracle}{HTML}{E15759}
\definecolor{color_proposed}{HTML}{F28E2B}

\definecolor{alg_gray}{HTML}{E8EEEB}
\definecolor{alg_blue}{HTML}{C5BEDF}
\definecolor{alg_yellow}{HTML}{FDF9C0}

\DeclareRobustCommand{\legmark}[3]{
  \tikz[baseline=-0.6ex]{
    \def\L{0.5cm}
    \draw[line width=2.0pt, draw=#1, #2] (0,0) -- (\L,0);
    \node[#3] at (0.25cm,0) {};
  }%
}

\tikzset{
alg_box/.style={
    draw=black, draw opacity=0.7, line width=1.7pt, rounded corners=2pt,
    align=center, minimum height=6mm, minimum width=35mm, inner sep=2.2pt
  },
alg_arrow/.style={
-{Stealth[length=2mm,width=2.1mm]}, draw=black, draw opacity=0.7, line width=1.8pt
}
}

\newcommand{\tightparagraph}[2][2.25mm]{%
  \vspace{-#1}%
  \paragraph{#2}%
}

\hypersetup{colorlinks=true, citecolor=linkcolor, linkcolor=linkcolor, urlcolor=linkcolor}

\newif\ifshownotes
\shownotestrue
\newcounter{note}

\title{When to Transfer: Adaptive Source Selection for Positive Transfer in Linear Models}

\author[1]{\href{mailto:<hamza.cherkaoui@telecom-sudparis.eu>?Subject=Your UAI 2026 paper}{Hamza Cherkaoui}{}}
\author[1,2]{Hélène~Halconruy}
\author[1]{Yohan~Petetin}

\affil[1]{%
    SAMOVAR\\
    Télécom SudParis\\
    Institut Polytechnique de Paris\\
    91120 Palaiseau
}
\affil[2]{%
    Modal'X\\
    Université Paris-Nanterre\\
    92000 Nanterre
}

\begin{document}
\maketitle

\begin{abstract}
    In many business settings, task-specific labeled data are scarce or costly to obtain, limiting supervised learning on a target task.
    A classical response is transfer learning (TL).
    Many TL works study \emph{how} to transfer information from related sources.
    We study, for linear regression and classification, \emph{when} to transfer via \emph{sample sharing}: in a multi-source setting, we greedily decide from which sources and how many samples to incorporate into the target dataset.
    Our method uses an accept/reject rule based on a data-dependent estimate of the \emph{transfer gain}, \ie the marginal decrease in target predictive error, computed \emph{conditionally} on the observed target samples.
    We analyze our approach and show that how the derived statistical test enforces positive transfer with high probability.
    Under additional standard conditions, we also study the transfer gain itself and characterize when transfer is beneficial.
    Experiments on synthetic and real data show consistent gains over classical and recent strong baselines while avoiding negative transfer.
\end{abstract}

\section{Introduction}
\label{sec:intro}

Prediction tasks are ubiquitous in data analysis~\citep{Phillips2005,HastieTibshiraniFriedman2009,ThomasEdelmanCrook2017,HyndmanAthanasopoulos2021}, with applications in engineering, finance, and healthcare.
In many business settings, task-specific labeled data are scarce, limiting supervised learning on a target task~\citep{Settles2009,PanYang2010}.

A classical solution is \emph{Transfer Learning}: leveraging information from another related source tasks to improve prediction on the target task.
Previous works studie \textbf{how} to transfer, including domain adaptation~\citep{BenDavid2010,Ganin2016}, fine-tuning~\citep{Yosinski2014}, distillation~\citep{Hinton2015}, and regularization- or prior-based methods~\citep{EvgeniouPontil2004,Li2018L2SP}.

However, equally important questions are \textbf{when} transfer helps, and how to design procedures that \textbf{guarantee improved performance} on the target task.
To obtain such guarantees, \citet{wang2019characterizing} introduce the notion of \emph{transfer gain} that quantifies the benefit of transferring from a source to a target task.
A key objective is then to enforce a positive gain, \resp to stop transferring when the source is no longer helpful.
In this line of work, \citet{obst2021transfer} reuse this formalism and control the transfer through the number of gradient-descent iterations that move the target-trained parameters toward the source parameters.
However, most of the approaches do not provide an explicit fine-grained control of the transfer gain to ensure a positive transfer.

In this paper, we \emph{explicitly} control transfer through \emph{sample sharing}: in a multi-source setting, we greedily decide from which sources and how many samples to incorporate into the target dataset.
We focus on ridge classification and regression, where beneficial transfer induces an explicit bias--variance trade-off that admits a closed-form characterization, enabling a clearer analysis and discussion.
\begin{figure}[ht]
  \centering
  \begin{minipage}{0.44\linewidth}
    \centering
    \setlength{\unitlength}{1pt}
    \begin{picture}(0,0)
      \put(-10,20){\rotatebox{90}{\small\emph{Test error}}}
      \put(5,-10){\footnotesize\emph{Number of target samples}}
    \end{picture}%
    \includegraphics[width=0.9\linewidth, trim=0.2cm 0.7cm 0.2cm 1.9cm, clip]{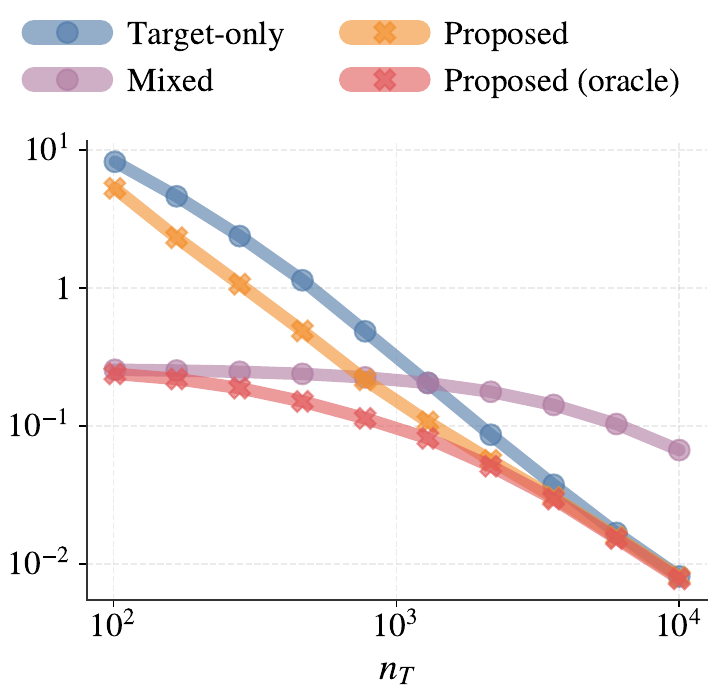}  
  \end{minipage}\hfill%
  \begin{minipage}{0.56\linewidth}
    \caption{\small Test error \vs Number of target samples for
      \legmark{color_all_target}{solid}{circle, draw=color_all_target, fill=color_all_target, minimum size=4.0pt, inner sep=1.2pt}\ \textit{All target samples},\;
      \legmark{color_all_target_source}{solid}{circle, draw=color_all_target_source, fill=color_all_target_source, minimum size=4.0pt, inner sep=1.2pt}\ \textit{All target + source samples},\;
      \legmark{color_oracle}{solid}{cross out, draw=color_oracle, line width=2.0pt, minimum size=4.0pt, inner sep=0pt}\ \textit{Proposed-Oracle},\;
      \legmark{color_proposed}{solid}{cross out, draw=color_proposed, line width=2.0pt, minimum size=4.0pt, inner sep=0pt}\ \textit{Proposed}.%
    }
    \label{FIG:SAMPLE_SHARING_INTRO}  
  \end{minipage}
\end{figure}

As illustrated in~\autoref{FIG:SAMPLE_SHARING_INTRO}\footnote{The experimental setting of this illustration is described in Appendix~\autoref{sec:supp_exp}.}, our approach provides conditions and procedures to enforce positive transfer, improving performance in the low-label regime while avoiding negative transfer (see the red and orange curves in~\autoref{FIG:SAMPLE_SHARING_INTRO}).

We summarize our contributions as follows:
\begin{enumerate}[leftmargin=2pt,itemsep=1pt,topsep=1pt,parsep=0pt,partopsep=0pt]
  \item \emph{Conditional transfer-gain criterion}: we introduce a target-conditioned transfer-gain statistic and derive decision rules that enforce positive transfer with high probability, for both regression and classification.
  \item \emph{Multi-source selection}: we propose an online greedy chunk-selection scheme that iteratively chooses \emph{when}, \emph{how much}, and \emph{from which} source to borrow.
  \item \emph{Theoretical characterization}: under additional assumptions, we characterize the transfer gain and discuss its finite-sample and asymptotic behavior.
  \item \emph{Empirical validation}: experiments on synthetic and real datasets confirm consistent improvements over over classical and recent strong baselines strong baselines while avoiding negative transfer.
\end{enumerate}

\section{Related Work}
\label{sec:related_work}

Our approach combines two main ingredients: (i) focusing on \emph{when} to transfer via sample sharing, and (ii) restricting to a linear setting to obtain a clean characterization.
Accordingly, we first review Transfer Learning (TL) broadly, then focus on linear transfer, before discussing work that enforces positive transfer, and finally mention sample-utility approaches, which are closely related to our sample-sharing perspective.


\paragraph{TL: General Overview} TL aims to improve a target predictor by reusing information learned from \emph{related} sources, most often to mitigate the limited availability of labeled target data~\citep{PanYang2010,WeissKhoshgoftaarWang2016,zhuang2020comprehensive,tan2018survey}.
TL can be implemented through several complementary mechanisms: adapting across domains when distributions differ but the prediction problem is aligned (domain adaptation)~\citep{BenDavid2010,Csurka2017,Ganin2016}; transferring parameters through pretraining--fine-tuning when large-scale source data are available~\citep{Yosinski2014}; compressing or guiding the target model via knowledge distillation~\citep{Hinton2015}; and encoding source information as inductive bias through shared priors or regularization, including classical multi-task formulations~\citep{EvgeniouPontil2004,ArgyriouEvgeniouPontil2007,Li2018L2SP}.
Beyond these approaches, practical pipelines often combine distributional corrections (\eg importance weighting under covariate shift)~\citep{Shimodaira2000,SugiyamaKrauledatMueller2007} with feature--distribution alignment methods (including optimal-transport-based matching)~\citep{CourtyFlamaryTuiaRakotomamonjy2017}, and multi-source settings further require aggregating heterogeneous sources, \eg via mixture-based strategies~\citep{MansourMohriRostamizadeh2009}.

A central difficulty is that task relatedness is often imperfect: transferring from a mismatched source can hurt, leading to \emph{negative transfer}~\citep{Rosenstein2005Transfer,zhang2022survey}.
This has motivated work on \emph{when} TL helps, including transferability measures and risk-based viewpoints~\citep{transferRisk2023}.


\paragraph{TL as Regularization for Linear Models} In linear prediction, TL is often realized through regularization that biases the target parameters toward source information.
Classical formulations couple source and target via penalty terms~\citep{EvgeniouPontil2004}, while hypothesis transfer methods fix a source estimator and shrink the target toward it, as in data-enriched regression~\citep{ChenOwenShi2014}.
Multi-task formulations extend this idea by enforcing shared structure, \eg via group sparsity to align supports~\citep{ObozinskiWainwrightJordan2010} or low-rank constraints to induce a common subspace~\citep{ArgyriouEvgeniouPontil2007}; domain adaptation techniques such as feature augmentation offer an equivalent linear view~\citep{Daume2007}.
Other strategies include stability-based methods that control the effect of source samples~\citep{kuzborskij2013stability} and adaptive algorithms guided by Bayesian optimization~\citep{sorocky2020share}.

In these approaches, coupling hyperparameters are typically tuned on held-out data, so success depends on source--target similarity; when this fails, negative transfer may occur~\citep{wang2019characterizing}.


\paragraph{TL with Positive Guarantees} Several works aim to \emph{enforce} positive transfer, either by filtering harmful source information or by certifying beneficial transfer under explicit conditions.
For instance, \citet{wang2019characterizing} characterize negative transfer and propose strategies to avoid it by identifying and down-weighting unrelated source data.
In the linear setting, \citet{obst2021transfer} formalize a transfer-gain criterion and provide guarantees when this gain is positive; in their setup, the amount of transfer is adjusted indirectly through the number of gradient-descent steps that move the target-trained parameters toward the source-trained parameters.
More recently, approaches have also considered controlling \emph{how much} to transfer by optimizing the quantity of transferred source data~\citep{otqms2025} or by deciding whether to merge datasets based on guaranteed population-loss reduction~\citep{joindisjoin2025}.

In contrast, we control transfer through \emph{sample sharing} and derive acceptance rules with positive-transfer guarantees for ridge regression, yielding a transparent bias--variance characterization \wrt the number of shared samples.


\paragraph{Selective Sample Sharing Across Tasks} Our sample-sharing perspective relates to selective data acquisition and subset selection, including active learning~\citep{Settles2009} and coreset methods~\citep{SenerSavarese2018}, which aim to improve learning by choosing which samples to label or retain.
It also connects to distribution-shift corrections and reweighting techniques (\eg covariate shift and importance weighting)~\citep{SugiyamaKawanabe2012}.

Unlike these approaches, which typically rank or weight samples within a single task, we study \emph{cross-task} sample sharing.

\section{Preliminaries}
\label{sec:single_case}

\tightparagraph{Notation}
We use lowercase (\eg $\alpha$) to denote a scalar, bold lowercase (\eg $\bm{x}$) to denote a vector, and bold uppercase (\eg $\bm{A}$) to denote a matrix.
The $\ell_2$-norm of a vector $\bm{x}$ is $\|\bm{x}\|_2=\sqrt{\bm{x}^\top\bm{x}}$.
For vectors of the same dimension, $\bm{a}\odot \bm{b}$ denotes the Hadamard (elementwise) product and $\bm{a}\oslash \bm{b}$ the elementwise division.
For a vector-valued function $\varphi$, $\varphi(\bm{x})$ is applied elementwise.

\tightparagraph{Data modeling}

In this section, we study the estimation of the unknown linear parameter common of both the regression and binary classification task. We consider a \emph{target} task with the training data set $\bm{X}_T\in\mathbb{R}^{n_T \times d}$ and $\bm{y}_T\in\mathbb{R}^{n_T}$ (\resp $\bm{y}_T\in\{-1, 1\}^{n_T}$ for the classification) given by
\[
    \bm{y}_T=\bm{X}_T\bm{\theta}_T^\star+\bm{\eta}_T \qquad (\text{\resp} \; \bm{y}_T=\mathrm{sign}(\bm{X}_T\bm{\theta}_T^\star+\bm{\eta}_T)) \;,
\]
where
$\bm{\theta}_T^\star\in\mathbb{R}^d$ is unknown, $\bm{X}_T, \bm{y}_T$ are observed, and $\bm{\eta}_T\in\mathbb{R}^{n_T}$ denotes noise. We assume that \( \mathbb{E}[\bm{\eta}_T]=\bm{0}_{n_T}, \quad \cov(\bm{\eta}_T)=\sigma_T^2 \I_{n_T} \) .

We also assume access to an independent validation data set \(\bm{X}_T^{\mathrm{val}}\in\mathbb{R}^{n_T^{\mathrm{val}}\times d}\) and \(\bm{y}_T^{\mathrm{val}}\in\mathbb{R}^{n_T^{\mathrm{val}}}\) (\resp $\bm{y}_T^{\mathrm{val}}\in\{-1, 1\}^{n_T}$ for the classification) drawn from the same distribution, with the feature matrix \(\bm{X}_T^{\mathrm{val}}\) observed.

\tightparagraph{Target-only estimation.}
To estimate $\bm{\theta}_T^\star$, we use a ridge regression with regularization parameter $\lambda_T \geq 0$ (reducing to the OLS when $\lambda_T=0$), defined as:
\[
    \wh{\bm{\theta}}_T := \arg\min_{\bm{\theta} \in \mathbb{R}^d} ~ \frac{1}{2}\big\|\bm{X}_T\bm{\theta} - \bm{y}_T\big\|_2^2 +\frac{\lambda_T}{2} \big\|\bm{\theta}\big\|_2^2 \;.
\]
By the first-order optimality condition, we obtain the following closed form
\begin{equation}\label{eq:theta_T}
    \wh{\bm{\theta}}_T = \bm{A}_T^{-1} \big( G_T \bm{\theta}_T^\star + \bm{Z}_T \big),
\end{equation}
with \(
\bm{A}_T = \bm{G}_T + \lambda_T \I_d \;\in\mathbb{R}^{d \times d} \;,\;
\bm{G}_T = {\bm{X}_T}^\top\bm{X}_T \;\in\mathbb{R}^{d \times d} \;,\;
\bm{Z}_T = {\bm{X}_T}^\top\bm{\eta}_T \;\in\mathbb{R}^{d}
\) .

\tightparagraph{Collaborative estimation.}

We study the estimation problem on the pooled dataset obtained by stacking the target data with the source data. We assume access to $n$ auxiliary \emph{source} samples with $0<n\le n_{\max}$, where $n_{\max}$ is a user-set budget (fixed \emph{a priori}). Consider the data set $\bm{X}_S\in\mathbb{R}^{n \times d}$ and \(\bm{y}_S\in\mathbb{R}^{n}\) (\resp $\bm{y}_S\in\{-1, 1\}^n$ for the classification) given by
\[
    \bm{y}_S=\bm{X}_S\bm{\theta}_S^\star+\bm{\eta}_S, \qquad (\text{\resp} \; \bm{y}_S=\mathrm{sign}(\bm{X}_S\bm{\theta}_S^\star+\bm{\eta}_S)) \;,
\]
where $\bm{\theta}_S^\star\in\mathbb{R}^d$ is unknown, $\bm{X}_S\in\mathbb{R}^{n \times d}$ are observed, and $\bm{\eta}_S$ denotes noise. We assume that \( \mathbb{E}[\bm{\eta}_S]=\bm{0}_{n}, \quad \cov(\bm{\eta}_S)=\sigma_S^2 \I_{n} \).

Let \(\wh{\bm{\theta}}(n)\) denote the \emph{collaborative ridge} estimator defined from the ridge objective formed by the \textbf{stacked} \emph{target} samples and the first \(n\) samples from the \emph{source} data set, \ie :
\begin{align*}
    \wh{\bm{\theta}}(n) & := \arg\min_{\bm{\theta} \in \mathbb{R}^d} ~ \frac{1}{2}\big\|\bm{X}_c\bm{\theta} - \bm{y}_c\big\|_2^2  +\frac{\lambda_c}{2} \big\|\bm{\theta}\big\|_2^2 \\
    \mathrm{with} \;    & \bm{X}_c^\top := \big[ \bm{X}_T^\top \; \bm{X}_S(n)^\top \big]^\top,\; \bm{Y}_c^\top := \big[ \bm{y}_T^\top \; \bm{y}_S(n)^\top \big]^\top               \\
    \bm{X}_S(n)         & := \Big[ \bm{x}_{S,1}^\top \;\dots\; \bm{x}_{S,n}^\top \Big]^\top,\; \bm{Y}_S(n) :=  \Big[ \bm{y}_{S,1} \;\dots\; \bm{y}_{S,n} \Big]^\top \;.
\end{align*}

The intuition is that when $\bm{\theta}_S^\star \approx \bm{\theta}_T^\star$, pooling datasets reduces variance and thus the target prediction error. By the first-order optimality condition, we obtain the following closed form
\begin{equation}
    \wh{\bm{\theta}}(n) = \bm{A}_c(n)^{-1} \big(\bm{G}_S(n) \bm{\theta}_S^\star + \bm{G}_T \bm{\theta}_T^\star + \bm{Z}_S(n)+ \bm{Z}_T \big),
\end{equation}
with \(
\bm{A}_c(n)   = \bm{G}_T + \bm{G}_S(n) + \lambda_c \I_d \;\in\mathbb{R}^{d \times d} \;,\;
\bm{G}_S(n)   = {\bm{X}_S(n)}^\top\bm{X}_S(n) \;\in\mathbb{R}^{d \times d} \;,\;
\bm{Z}_S(n)   = {\bm{X}_S(n)}^\top\bm{\eta}_S(n) \;\in\mathbb{R}^{d} \;,\;
\bm{\eta}_S(n) =  \Big[ \eta_{S,1} \;\dots\; \eta_{S,n} \Big]^\top
\) .

\section{Prediction tasks}
\label{sec:predicting}

We exploit the previous estimation steps for prediction. In regression, given a feature vector \(\bm{x}\), we produce the prediction \(\hat{y}\) as
\mbox{\( \hat{y}=\bm{x}^\top \wh{\bm{\theta}}_T \quad (\resp\ \hat{y}=\bm{x}^\top \wh{\bm{\theta}}(n))\)},
while in binary classification we consider \mbox{\( \hat{y}=\sign\!\big(\bm{x}^\top \wh{\bm{\theta}}_T\big) \quad (\resp\ \hat{y}=\sign\!\big(\bm{x}^\top \wh{\bm{\theta}}(n)\big)) \)}.

As we aim to enforce positive transfer, we next define prediction errors that quantify performance in each prediction task.
First, we consider the target-only estimator, then the collaborative estimator built from target and source samples for both regression and classification task.
We evaluate performance on the validation set $(\bm{X}_T^{\mathrm{val}},\bm{y}_T^{\mathrm{val}})$ and, unless stated otherwise, we condition on the design matrices $\bm{X}_T$, $\bm{X}_T^{\mathrm{val}}$, and $\bm{X}_S$, taking expectations with respect to observation \textbf{noise only}, to avoid assumptions on the feature distribution.

\tightparagraph{Target-only regression}
We define the target regression error on the validation set as:
{\small
\begin{align*}
    \xi_T
     & := \mathbb{E}\!\left[ \left\| \bm{X}_T^{\mathrm{val}} (\wh{\bm{\theta}}_{T} \,-\, \bm{\theta}_T^\star) \right\|_2^2 \right]           \\
     & = \lambda_T^2\left\| \bm{U}_T \bm{\theta}_T^\star \right\|_2^2 \,+\, \sigma_T^2\,\tr\!\big(\bm{U}_T \bm{G}_T {\bm{U}_T}^\top\big) \;,
\end{align*}
}
with \( \bm{U}_T \,:=\, \bm{X}_T^{\mathrm{val}}\bm{A}_T^{-1} \) .
This error is composed of two components: a shrinkage bias and a noise variance.

\tightparagraph{Collaborative regression}
We define the collaborative regression error on the validation set as:
{\small
\begin{align*}
    \xi(n)
     & := \mathbb{E}\! \left[ \left\|  \bm{X}_T^{\mathrm{val}} (\wh{\bm{\theta}}(n) - \bm{\theta}_T^\star \big) \right\|_2^2 \right]           \\
     & = \left\| \bm{V}(n) \!\left(\bm{G}_S(n)\,(\bm{\theta}_S^\star - \bm{\theta}_T^\star)-\lambda_c\,\bm{\theta}_T^\star\right) \right\|_2^2 \\
     & \,+\, \tr\!\Big(\bm{V}(n) \big( \sigma_T^2 \bm{G}_T + \sigma_S^2 \bm{G}_S \big) {\bm{V}(n)}^\top \Big)       \;,
\end{align*}
}
with \( \bm{V}(n) \,:=\, \bm{X}_T^{\mathrm{val}} \bm{A}_c(n)^{-1} \) .
The error is composed of a first approximation and shrinkage bias term and a second noise term; the latter combines source and target noise through the collaborative Gram matrix.

\tightparagraph{Target-only classification}
We now consider binary classification.
Let $\bm{1}\in\mathbb{R}^{n_T^{\mathrm{val}}}$ denote the all-ones vector.
While the $01$-loss is a natural choice, it is discontinuous.
Hence to obtain an analyzable error, we use the probit surrogate \( \Phi(u) := \int_{-\infty}^{t} \frac{1}{\sqrt{2\pi}}\,\exp\!\left(-\frac{u^{2}}{2}\right)\,du \), which can be interpreted as the misclassification probability under a Gaussian perturbation of the margin~\citet{Bishop2006PRML}.
We define the target classification error on the validation set as:
{\small
\begin{align*}
    \zeta_T
    :=\, & \frac{1}{n_T^{\mathrm{val}}}\,\mathbb{E}\!\left[\bm{1}^\top \Phi\!\Big( -\,\bm{y}_T^{\mathrm{val}}\odot\big(\bm{X}_T^{\mathrm{val}}\wh{\bm{\theta}}_T\big) \Big) \right] \\
    =\,  & \frac{1}{n_T^{\mathrm{val}}}\, \bm{1}^\top\Big(\bm{\pi}\odot \Phi(-\bm{t}_T)\;+\;(\bm{1}-\bm{\pi})\odot \Phi(\bm{t}_T) \Big),
\end{align*}
}
where $\bm{\pi}:=\mathbb{P}(\bm{y}_T^{\mathrm{val}}=\bm{1}\mid \bm{X}_T^{\mathrm{val}})$ and where we defined
\mbox{\small \( \bm{t}_T \;:=\; (\bm{U}_T\bm{g}_T)\oslash \sqrt{\bm{1}+\bm{s}_T^{2}}\)},
\mbox{\small \( \bm{U}_T := \bm{X}_T^{\mathrm{val}}\bm{A}_T^{-1}\)},
\mbox{\small \( \bm{g}_T := \bm{G}_T\bm{\theta}_T^\star\)},
\mbox{\small \( \bm{s}_T^{2} := \sigma_T^2\,\diag\!\big(\bm{U}_T \bm{G}_T {\bm{U}_T}^\top \big)\)}.
The vector $\bm{t}_T$ is a per-point signal-to-uncertainty margin, and $\bm{s}_T^{2}$ is the per-point predictive variance induced by the noise.

\tightparagraph{Collaborative classification}
We define the collaborative classification error on the validation set as:
{\small
\begin{align*}
    \zeta(n)
    :=\, & \frac{1}{n_T^{\mathrm{val}}}\,\mathbb{E}\!\left[\bm{1}^\top \Phi\!\Big( -\,\bm{y}_T^{\mathrm{val}}\odot\big(\bm{X}_T^{\mathrm{val}}\wh{\bm{\theta}}(n)\big) \Big) \right] \\
    =\,  & \frac{1}{n_T^{\mathrm{val}}}\, \bm{1}^\top\Big(\bm{\pi}\odot \Phi(-\bm{t}_c(n))\;+\;(\bm{1}-\bm{\pi})\odot \Phi(\bm{t}_c(n)) \Big),
\end{align*}
}
where we defined
\mbox{\small \( \bm{t}_c(n) \;:=\; \big(\bm{V}(n)\bm{g}_c(n)\big)\oslash \sqrt{\bm{1}+\bm{s}_c^{2}(n)}\)},
\mbox{\small \( \bm{V}(n) := \bm{X}_T^{\mathrm{val}}\bm{A}_c(n)^{-1}\)},
\mbox{\small \( \bm{g}_c(n) := \bm{G}_T\bm{\theta}_T^\star + \bm{G}_S(n)\bm{\theta}_S^\star\)},
\mbox{\small \( \bm{s}_T^{2}(n) := \sigma_T^2\,\diag\!\big(\bm{V}(n)\bm{G}_T{\bm{V}(n)}^\top \big)\)},
\mbox{\small \( \bm{s}_S^{2}(n) := \sigma_S^2\,\diag\!\big(\bm{V}(n)\bm{G}_S(n){\bm{V}(n)}^\top \big)\)},
\mbox{\small \( \bm{s}_c^{2}(n) := \bm{s}_T^{2}(n)+\bm{s}_S^{2}(n)\)}.
The vector $\bm{t}_c(n)$ is a per-point signal-to-uncertainty margin for the collaborative predictor, and
$\bm{s}_c^{2}(n)$ is the per-point predictive variance induced by the combined target/source noise through $\wh{\bm{\theta}}(n)$.

All computation details of this section are reported in Appendix~\autoref{sec:supp_est_errors}.

\section{Positive transfer}
\label{sec:collab_case}

Now that we have define the prediction tasks and quantify their performance in both the task-only and collaborative setting, a natural question is: \emph{Which setting (single-task or collaborative) achieves better prediction?} To answer it, we define the \emph{transfer gain}, in the regression and classification cases, which measures the reduction in prediction error when moving from a single task to collaborative training.

\begin{restatable}[Transfer gain]{definition}{defa}\label{def:transfert_grain}
    We define the \emph{transfer gain} criterion as the reduction in prediction error due to sharing \ie \( \Delta_{\mathrm{regr}}^\star(n) := \xi_T - \xi(n) \) \; (\resp \( \Delta_{\mathrm{clf}}^\star(n) := \zeta_T - \zeta(n) \)) .
\end{restatable}

Our definition of transfer gain coincides, for the regression case, with that of \citet{obst2021transfer}, who formalize it as the difference in quadratic prediction error (QPE; see \citet{bosq2008inference}) between an estimator trained solely on the target sample and a fine-tuned estimator. In contrast, the notion of a negative transfer gap introduced by \citet{wang2019characterizing} quantifies detrimental transfer: it is negative whenever the expected risk (with respect to any loss function) of an algorithm using both source and target data exceeds that of an algorithm trained exclusively on the target data.

\begin{restatable}[Positive Transfer gain]{definition}{defb}\label{def:beneficial_collaboration}
    Transfer gain is \emph{positive} if $\Delta_{\mathrm{regr}}^\star(n) \ge 0$ (\resp $\Delta_{\mathrm{clf}}^\star(n) \ge 0$) and \emph{negative} if \( \Delta_{\mathrm{clf}}^\star(n) < 0 \) (\resp \( \Delta_{\mathrm{regr}}^\star(n) < 0 \)).
\end{restatable}

A positive value (\resp negative) indicates an improvement (\resp degradation) compared to training only on the target. Thus, the oracle decision is: \emph{borrow samples if \( \Delta_{\mathrm{regr}}^\star(n) \ge 0 \) (\resp \( \Delta_{\mathrm{clf}}^\star(n) \ge 0 \) ), otherwise abstain}.

\subsection{Estimating the transfer gain}

A limitation of the transfer gains $\Delta_{\mathrm{regr}}^\star(n)$ and $\Delta_{\mathrm{clf}}^\star(n)$ is that they depend on the unknown parameters $\bm{\theta}_T^\star$ and $\bm{\theta}_S^\star$.
To make the criterions operational, we derive a plug-in estimators, denoted $\wh{\Delta}_{\mathrm{regr}}(n)$ and $\wh{\Delta}_{\mathrm{clf}}(n)$.

\begin{restatable}[Estimated transfer gain for regression]{definition}{defc}\label{def:estimated_transfert_grain_regr}
    We denote by $\wh{\Delta}_{\mathrm{regr}}(n)$ the plug-in estimator of the transfer gain $\Delta_{\mathrm{regr}}^\star(n)$.
        {\small
            \begin{align*}
                \wh{\Delta}_{\mathrm{regr}}(n)
                                                      & := \lambda_T^2 \| \bm{U}_T \wh{\bm{\theta}}_T \|_2^2                                                                  \\
                \; \,-\,                    \| \bm{V} & (n) \big( \bm{G}_S(n) \big( \wh{\bm{\theta}}_S - \wh{\bm{\theta}}_T \big) - \lambda_c \wh{\bm{\theta}}_T \big) \|_2^2 \\
                \; \,+\,                    \tr \big( & \bm{U}_T \bm{M}(n) {\bm{U}_T}^{\top} \big) \,-\, \tr \big( \bm{V}(n) \bm{N}(n) {\bm{V}(n)}^{\top} \big)               \\
                \text{where} \;    \bm{M}_T           & := \sigma_T^2 \bm{G}_T - \lambda_T^2 \sigma_T^2 {\bm{A}_T}^{-1} \bm{G}_T {\bm{A}_T}^{-1}                              \\
                \bm{N}(n)                             & := \bm{K}_1(n) + \bm{K}_2(n) + \sigma_S^2 \bm{G}_S(n) + \sigma_T^2 \bm{G}_T \,,                                       \\
                \bm{K}_1(n)                           & := - \sigma_S^2 \bm{G}_S(n) {\bm{A}_S(n)}^{-1} \bm{G}_S(n) {\bm{A}_S(n)}^{-1} \,,                                     \\
                \bm{K}_2(n)                           & := - \sigma_T^2 \bm{A}_S(n) {\bm{A}_T}^{-1} \bm{G}_T {\bm{A}_T}^{-1} \bm{A}_S(n) \,.
            \end{align*}
        }
\end{restatable}

And we do similarly for the classification case.

\begin{restatable}[Estimated transfer gain for classification]{definition}{defd}\label{def:estimated_transfert_grain_clf}
    We denote by $\wh{\Delta}_{\mathrm{regr}}(n)$ the plug-in estimator of the transfer gain $\Delta_{\mathrm{regr}}^\star(n)$.
        {\small
            \begin{align*}
                \wh{\Delta}_{\mathrm{clf}}(n)
                =\, & \frac{1}{n_T^{\mathrm{val}}}\, \bm{1}^\top\Big( \bm{\pi}\odot\big[\Phi(-\wh{\bm{t}}_T)-\Phi(-\wh{\bm{t}}_c(n)) \big] \Big)            \\
                +\, & \frac{1}{n_T^{\mathrm{val}}}\, \bm{1}^\top\Big( (\bm{1}-\bm{\pi})\odot\big[\Phi(\wh{\bm{t}}_T)-\Phi(\wh{\bm{t}}_c(n)) \big] \Big) \,.
            \end{align*}
        }
\end{restatable}

To ensure that \( \wh{\Delta}_{\mathrm{regr}}(n) \) and \( \wh{\Delta}_{\mathrm{clf}}(n) \) are reasonable in practice, we compare it to the empirical counterpart with $\varepsilon = \| \bm{\theta}_T^\star - \bm{\theta}_S^\star \|_2 = 0.2$.

\begin{figure}[t]
    \centering
    \begin{tabular}{cc}
        \textit{a. Regression}                                                           & \textit{b. Classification} \\
        \includegraphics[width=0.21\textwidth]{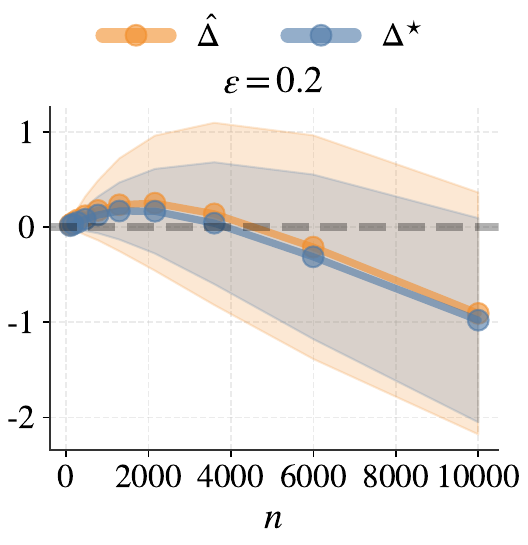} &
        \includegraphics[width=0.21\textwidth]{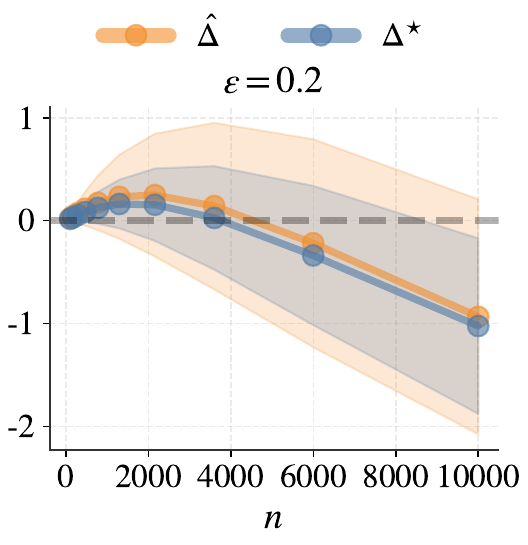}                           \\
    \end{tabular}
    \caption{Empirical $\Delta_{\mathrm{regr}}^\star(n)$ and estimate $\wh{\Delta}_{\mathrm{regr}}(n)$ (left), and their classification counterparts $\Delta_{\mathrm{clf}}^\star(n)$ and $\wh{\Delta}_{\mathrm{clf}}(n)$ (right); the estimator captures the beneficial-transfer region.}
    \label{FIG:DELTA_ILLUS_EPS_0_2}
\end{figure}

In~\autoref{FIG:DELTA_ILLUS_EPS_0_2} (left), we observe that the estimators closely tracks the empirical curve and correctly identifies the range where transfer is beneficial.
Next, we characterize the estimator by deriving its expectation and variance.

\subsection{Characterizing the gain}

\begin{restatable}[Expectation and variance of $\wh{\Delta}_{\mathrm{regr}}(n)$]{property}{propa}\label{prop:expect_var_delta_hat_regr}
    Considering the plug-in estimator $\wh{\Delta}_{\mathrm{regr}}(n)$, we have:
    {\small
    \begin{align*}
        \mathbb{E}\Big[ \wh{\Delta}_{\mathrm{regr}}(n) \Big] & = \Delta_{\mathrm{regr}}^\star(n) + b(n)                                       \\
        \var\Big[ \wh{\Delta}_{\mathrm{regr}}(n \big) \Big]  & = 2 \, \tr \big( (\bm{D}(n) \bm{\Sigma}(n))^2 \big) \,+\,                      \\
                                                             & \hspace{0.8cm} 4 \, \bm{\mu}(n)^\top \bm{D}(n) \bm{\Sigma}(n) \bm{D}(n) \mu(n)
    \end{align*}
    }
    Defining the bias and variance terms by:
    {\small
    \begin{align*}
        b(n) \, :=                                         & \, \lambda_T^4 \, \Big\| \bm{U}_T {\bm{A}_T}^{-1} \bm{\theta}_T^\star \Big\|_2^2                                                                     \\
        \quad                                - 2           & \, \lambda_T^2 \, \langle \bm{U}_T \bm{\theta}_T^\star,\; \lambda_T \bm{U}_T {\bm{A}_T}^{-1} \bm{\theta}_T^\star \rangle                             \\
        \quad                                - \Big\|      & \bm{V}(n) \Big( \bm{G}_S(n) \bm{\Delta\theta}^{\mathrm{b}} - \lambda_c \lambda_T {\bm{A}_T}^{-1} \bm{\theta}_T^\star \Big) \Big\|_2^2                \\
        \quad                                - \Big\langle & \bm{V}(n) \Big( \bm{G}_S(n) \bm{\Delta\theta} - \lambda_c \bm{\theta}_T^\star \Big) \,,                                                              \\
                                                           & \bm{V}(n) \Big( \bm{G}_S(n) \bm{\Delta\theta}^{\mathrm{b}}  + \lambda_c \lambda_T {\bm{A}_T}^{-1} \bm{\theta}_T^\star \Big) \Big\rangle              \\
        \bm{\Delta\theta} :=                               & \bm{\theta}_S^\star - \bm{\theta}_T^\star \,,\, \bm{W}(n) := \bm{V}(n)^\top \bm{V}(n)   \,,                                                          \\
        \bm{\Delta\theta}^{\mathrm{b}} :=                  & \lambda_T {\bm{A}_T}^{-1} \bm{\theta}_T^\star - \lambda_S {\bm{A}_S(n)}^{-1} \bm{\theta}_S^\star \,,                                                 \\
        \bm{D}(n) :=                                       & \begin{bmatrix}
                                                                 - \bm{G}_S(n) \bm{W}(n) \bm{G}_S(n) & \bm{G}_S(n) \bm{W}(n) \bm{A}_S(n) \\
                                                                 \bm{A}_S(n) \bm{W}(n) \bm{G}_S(n)   & \bm{D}_{22}(n)
                                                             \end{bmatrix} \,,                                                                              \\
        \bm{D}_{22}(n) =                                   & \lambda_T^2 \bm{U}_T^\top \bm{U}_T - \bm{A}_S(n) \bm{W}(n) \bm{A}_S(n) \,,                                                                           \\
        \bm{\mu}(n) :=                                     & \Big[ \Big( \bm{A}_S(n)^{-1} \bm{G}_S(n) \bm{\theta}_S^\star \Big)^\top \, \Big( {\bm{A}_T}^{-1} \bm{G}_T \bm{\theta}_T^\star \Big)^\top \Big]^\top, \\
        \bm{\Sigma}(n) :=                                  & \diag \big(\sigma_S^2 {\bm{A}_S(n)}^{-1} \bm{G}_S(n) {\bm{A}_S(n)}^{-1} \,,                                                                          \\
                                                           & \hspace{1.0cm}  \sigma_T^2 {\bm{A}_T}^{-1} \bm{G}_T {\bm{A}_T}^{-1} \big) \,.
    \end{align*}
    }
\end{restatable}

Note, that when $\lambda_T = \lambda_S = 0$, we have $\bm{\Delta\theta}^{\mathrm{b}} = 0$, which directly implies $ b(n)=0$. Consequently, we have:  \( \mathbb{E}\Big[ \wh{\Delta}_{\mathrm{regr}}(n) \Big] = \Delta_{\mathrm{regr}}^\star(n) \) in that case. We do the same characterization for the classification task.

\begin{restatable}[Expectation and variance of $\wh{\Delta}_{\mathrm{clf}}(n)$]{property}{propb}\label{prop:expect_var_delta_hat_clf}
    Assuming we know \( \bm{\pi} := \mathbb{P}(\bm{y}_T^{\mathrm{val}}=\bm{1}\mid \bm{X}_T^{\mathrm{val}})\in[0,1]^{n_T^{\mathrm{val}}} \), consider the plug-in estimator
        {\small
            \begin{align*}
                \mathbb{E}\!\left[\wh{\Delta}_{\mathrm{clf}}(n)\right]
                 & =\frac{1}{n_T^{\mathrm{val}}}\,\bm{1}^\top\,\mathbb{E}\!\left[\wh{\bm{m}}(n)\right],                     \\
                \var\!\left[\wh{\Delta}_{\mathrm{clf}}(n)\right]
                 & =\frac{1}{(n_T^{\mathrm{val}})^2}\Big(\mathbb{E}\!\left[\bm{1}^\top \wh{\bm{v}}_{\mathrm{val}}(n)\right]
                +\bm{1}^\top \cov\!\big(\wh{\bm{m}}(n)\big)\bm{1}\Big).
            \end{align*}
        }
    Defining the variance terms by:
    {\small
    \begin{align*}
        \wh{\bm{m}}(n) \,:=\;                & \bm{\pi}\odot\wh{\bm{d}}_+(n)+(\bm{1}-\bm{\pi})\odot\wh{\bm{d}}_-(n) \,,                                                \\
        \wh{\bm{v}}_{\mathrm{val}}(n) \,:=\, & \bm{\pi}\odot \wh{\bm{d}}_+(n)^{\odot 2}+(\bm{1}-\bm{\pi})\odot \wh{\bm{d}}_-(n)^{\odot 2}-\wh{\bm{m}}(n)^{\odot 2} \,, \\
        \text{where}\;\wh{\bm{r}}_T \,:=\,   & \bm{S}_T\bm{U}_T\wh{\bm{g}}_T,\qquad \wh{\bm{r}}_c \,:=\, \bm{S}_c(n)\bm{V}(n)\wh{\bm{g}}_c(n) \,,                      \\
        \wh{\bm{d}}_+(n) \,:=\,              & \Phi(-\wh{\bm{r}}_T)-\Phi(-\wh{\bm{r}}_c),\qquad \wh{\bm{d}}_-(n) \,:=\, \Phi(\wh{\bm{r}}_T)-\Phi(\wh{\bm{r}}_c) \,,    \\
        \bm{S}_T \,:=\,                      & \diag\!\big((\bm{1}+\bm{s}_T^2)^{-1/2}\big) \,,                                                                         \\
        \bm{S}_c(n) \,:=\,                   & \diag\!\big((\bm{1}+\bm{s}_c^2(n))^{-1/2}\big) \,,
    \end{align*}
    }
    with plug-in signals, for instance \( \wh{\bm{g}}_T:=\bm{G}_T\wh{\bm{\theta}}_T\) and \(\wh{\bm{g}}_c(n):=\bm{G}_T\wh{\bm{\theta}}_T+\bm{G}_S(n)\wh{\bm{\theta}}_S(n)\) .
\end{restatable}

The additional assumption of access to $\bm{\pi}$ is specific to classification; in practice, $\bm{\pi}$ can be estimated from the validation set.
Moreover, if the plug-in signals are exact (i.e., \(\wh{\bm{g}}_T=\bm{g}_T\) and \(\wh{\bm{g}}_c(n)=\bm{g}_c(n)\) a.s.), then \( \mathbb{E}[\wh{\Delta}_{\mathrm{clf}}(n)]=\Delta_{\mathrm{clf}}^\star(n) \) and the above variance reduces to the oracle one.

\subsection{Derivation of a pratical criterion}

\tightparagraph{Transfer grain lower bounds}

The previous estimators of the tansfer gain are random variable whose expectation and variance has be derived in closed form.
This allows us to establish a finite-sample lower bound.

\begin{restatable}[Transfer gain lower bounds]{property}{propc}\label{prop:lower_bounds_transfer_gain}
    We have with probability $ 0 < \delta < 1$:
    {\small
    \begin{align*}
        \Delta_{\mathrm{regr}}^\star(n)  < \,         & \mathrm{LCB}^{\mathrm{regr}}_\delta(n) \; \big(\text{\resp} \,  \Delta_{\mathrm{clf}}^\star(n)  < \, \mathrm{LCB}^{\mathrm{clf}}_\delta(n) \big) \\
        \text{with}    \hspace{15mm}                  &                                                                                                                                                  \\
        \mathrm{LCB}^{\mathrm{regr}}_\delta(n)  := \, & \wh{\Delta}_{\mathrm{regr}}(n) - b(n) - \sqrt{\var\big[ \wh{\Delta}_{\mathrm{regr}}(n) \big] \frac{1 - \delta}{\delta}}                          \\
        \mathrm{LCB}^{\mathrm{clf}}_\delta(n)  := \,  & \wh{\Delta}_{\mathrm{clf}}(n) - \sqrt{\var\big[ \wh{\Delta}_{\mathrm{clf}}(n) \big] \frac{1 - \delta}{\delta}}
    \end{align*}
    }
\end{restatable}

All computation details of this section are reported in Appendix~\autoref{sec:supp_prop_delta_hat}. All experimental setting details are available in Appendix~\autoref{sec:supp_exp}.

\tightparagraph{Positive tranfer test}

From the previous lower bounds, a natural appraoch is to design a \textbf{statistical tests} that, with high probability, detect when the true transfer gain is negative and therefore stop sample sharing.
We introduce the following one-sided hypotheses: \( H_0:\ \Delta_{\mathrm{regr}}^\star(n)\le 0 \quad\text{vs}\quad H_1:\  \Delta_{\mathrm{regr}}^\star(n)>0 \) (a similarly for the classification task) and define a valid test obtained by rejecting $H_0$ whenever \( \mathrm{LCB}^{\mathrm{regr}}_\delta(n) > 0, \) which certifies beneficial transfer with Type~I error at most $\delta$.

\tightparagraph{Positive tranfer practical criterion}

However, in practice, the test level $\delta$ can be overly conservative in finite samples; practitioners may want to tune the transfer to reflect their preferred trade-off between caution (avoiding negative transfer) and borrowing (seeking larger improvements).
We therefore derive from these lower bounds a UCB-inspired~\citep{AuerCesaBianchiFischer2002} \textbf{decision statistic} that selects how many source samples to borrow.
It is controlled by a single, easy-to-set parameter $\alpha$, which governs this caution--borrowing trade-off.
In~\autoref{sec:exp}, we show that setting $\alpha=0.01$ across \textbf{all} experiments consistently yields positive transfer and is simpler to tune in practice.

\begin{restatable}[Practical transfer gain]{definition}{defe}
    \label{def:conservative_transfert_grain}
    We define:
    {\small
    \begin{equation*}
        \textsc{TrGa}\big(\alpha, n \big) := \wh{\Delta}_{\mathrm{regr}}(n) - \alpha \sqrt{\wh{\var}\big( \wh{\Delta}_{\mathrm{regr}}(n) \big)}
    \end{equation*}
    }
    with $\wh{\var}\big( \wh{\Delta}_{\mathrm{regr}}(n) \big)$ the plug-in estimator of $\var\big( \wh{\Delta}_{\mathrm{regr}}(n) \big)$ defined using:
    {\small
    \begin{equation*}
        \wh{\bm{\mu}} = \Big[ \Big(\bm{A}_S(n)^{-1} \bm{G}_S(n) \wh{\bm{\theta}}_S\Big)^\top \; \Big({\bm{A}_T}^{-1} \bm{G}_T \wh{\bm{\theta}}_T\Big)^\top \Big]^\top \; .
    \end{equation*}
    }
\end{restatable}

It remains to efficiently evaluate multiple sizes~$n$ from multiple sources datasets.
All computation details of this section are reported in Appendix~\autoref{sec:supp_est_errors}.

\section{Proposed algorithm}
\label{sec:alg}

To efficiently exploit the transfer-gain statistic, we follow an online chunk-selection procedure over $S$ sources.

At each round, the algorithm considers the set of \emph{active} sources (those with remaining samples).
Depending on the strategy, it either (i) samples one active source uniformly, or (ii) evaluates a candidate chunk from \emph{each} active source and selects the chunk that maximizes the decision score (greedy).
The selected chunk is then appended to the borrowed set and the sufficient statistics are updated accordingly.
Efficient updates after admitting a chunk are performed via rank-one inverse updates (Sherman--Morrison/Woodbury).

The main steps are summarized in \autoref{fig:main_alg_flowchart}.
\begin{figure}[H]
  \centering
  \begin{tikzpicture}[font=\scriptsize, node distance=2.mm]
    \node[alg_box, fill=alg_yellow, fill opacity=0.7] (cond) {$n<n_{\max}$?};
    \node[alg_box, fill=alg_blue, fill opacity=0.7, below=of cond] (sources) {Loop over sources $s$};
    \node[alg_box, fill=alg_blue, fill opacity=0.7, below=of sources] (chunks) {Loop over samples from $s$, compute $\Delta(s,n_s)$};
    \node[alg_box, fill=alg_blue, fill opacity=0.7, below=of chunks] (accept) {$\wh{s}, \wh{n_s} \leftarrow \arg\max_{s, n_s} \Delta(s,n_s)$\\Accept $\wh{n_s}$ samples from $\wh{s}$, update $n$};

    \coordinate (Top) at ($(cond.north)+(0,3mm)$);
    \draw[alg_arrow] (Top) -- (cond.north);

    \draw[alg_arrow] (cond) -- (sources);
    \draw[alg_arrow] (sources) -- (chunks);
    \draw[alg_arrow] (chunks) -- (accept);

    \coordinate (L) at ($(cond.west)+(-8mm,0)$);
    \draw[alg_arrow] (accept.west) -- (L |- accept.west) -- (L |- cond.west) -- (cond.west);
\end{tikzpicture}
  \caption{Flowchart of the main steps of our approach.}
  \label{fig:main_alg_flowchart}
\end{figure}

A complete description of the algorithm and its complexity analysis are provided in Appendix~\autoref{sec:supp_alg}.

\section{Experiments}
\label{sec:exp}

In this section, we examine how the key problem parameters affect performance and show that our method improves target prediction on both synthetic and real datasets. All experiments were run in Python on a laptop-class CPU (Intel i7-7600U, 2 cores @ 2.80\,GHz). The code is publicly available at \href{https://github.com/***/***}{this repository}.

\subsection{Synthetic data benchmarks}
\label{subsec:simulation}

We first evaluate our approach on synthetic data. This section specifies the default settings that we adapt to each benchmark as needed. Unless stated otherwise, we generate \(10000\) target samples and \(10000\) source samples under the linear model specified in~\autoref{sec:single_case}. The ground-truth parameters \(\bm{\theta}_T, \bm{\theta}_S \in \mathbb{R}^{50}\) are random Gaussian vectors drawn at each run. We set \( \sigma_T=1 \) and \( \sigma_S=1 \) by default. The target ridge parameter \( \lambda_T \) is chosen by an oracle grid search on the target validation set \( \bm{X}_T^{\mathrm{val}}\). For the source, we fix \( \lambda_S=1 \) and use \( \lambda_{c}=\lambda_T+\lambda_S \). We set \(\alpha=0.01\) and \(n_{\max}=10000\) for all experiments. Each experiment is repeated \(250\) times, and we report averages over runs. Additional details and results are provided in Appendix~\autoref{sec:supp_exp}.

\tightparagraph{Effect of target sample size $n_T$:} We study how performance varies with the number of target samples by varying $n_T$ from $100$ to $10000$, while fixing the available source samples at $n_{\max}=10000$, \ie $\wh{n} \in \{0,\dots,10000\}$. We fix \(\varepsilon \in \{0.2, 8.0\}\) to study two \emph{extreme} contrasted scenarios (\ie $20\%$ and $80\%$ of the target linear parameter $\ell_2$-norm). All other settings follow the defaults. Performance is evaluated by the empirical test risk computed on held-out samples, \( \mathrm{err}(\bm{\theta}) := \big\| \bm{X}_T^{\mathrm{test}} \big(\bm{\theta}-\bm{\theta}_T^\star\big) \big\|_2^2 \;/\; n_T^{\mathrm{test}} \). We compare our proposed approach (and its Oracle version using the true $\bm{\theta}_T^\star$ and $\bm{\theta}_S^\star$) against (i) target-only ridge, (ii) mixed-task ridge, (iii) the data-enriched method of \citet{ChenOwenShi2014}, (iv) the approach of \citet{obst2021transfer}, (v) the approach of \citet{zhang2026unified}. Moreover, we include (vi) a hierarchical Bayesian ridge regression baseline that fits a Gaussian mixture model\footnote{We detail the model in Appendix~\autoref{sec:supp_exp}.} to illustrate how mixed-effects approaches behave in this setting.
\begin{figure}[t]
  \centering
  \includegraphics[width=.65\linewidth, trim=0.5cm 11.6cm 0.5cm 0.1cm, clip]{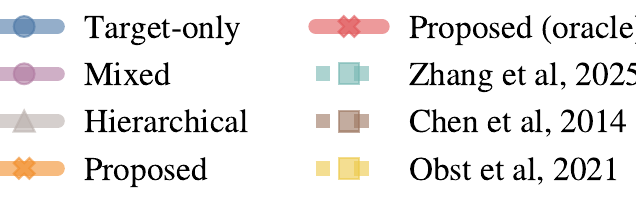}  
  \begin{subfigure}{.48\columnwidth}
    \centering
    \includegraphics[width=0.9\linewidth, trim=0.4cm 0.2cm 0.6cm 3.3cm, clip]{figures/04_performance_error_comparison__eps_0.2.pdf}  
    \subcaption{Low target--source discrepancy}\label{fig:exp_synth_eps_0.1}
  \end{subfigure}\hfill%
  \begin{subfigure}{.48\columnwidth}
    \centering
    \includegraphics[width=0.9\linewidth, trim=0.4cm 0.2cm 0.6cm 3.3cm, clip]{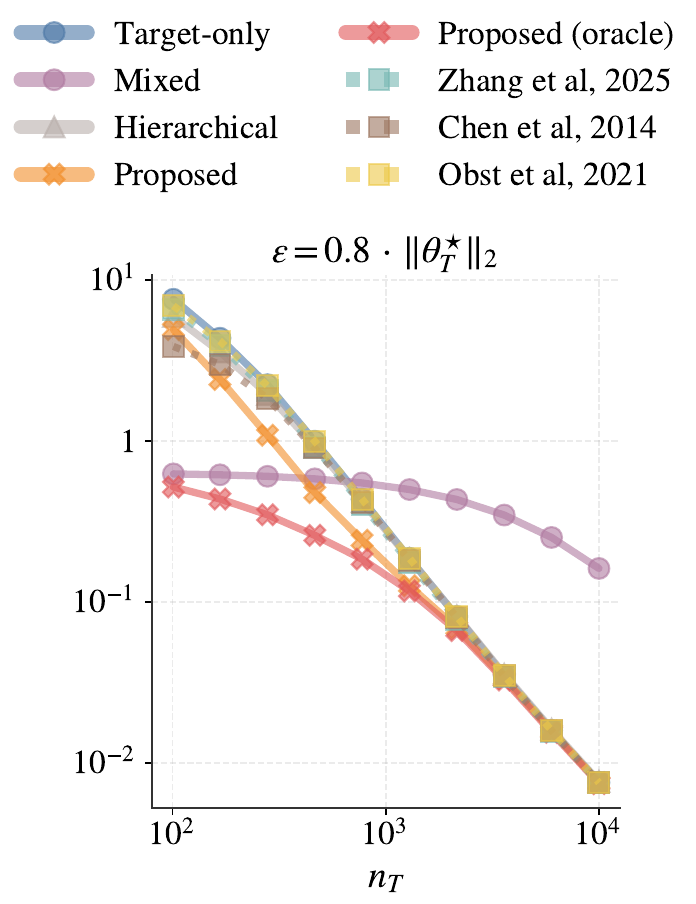}  
    \subcaption{Too high target--source discrepancy}\label{fig:exp_synth_eps_1.0}
  \end{subfigure}
  \caption{Predictive error comparison \wrt the number of target samples. The solid line reports the average.}
  \label{fig:synth_data_perf_exp}
\end{figure}
\autoref{fig:synth_data_perf_exp} reports the predictive error for target-only (solid blue), mixed-task (solid gray), \citet{obst2021transfer} (dashed yellow), \citet{ChenOwenShi2014} (dashed green), \citet{zhang2026unified} (dashed light gray), hierarchical (solid gray), our approach (solid orange) and its oracle counterpart (solid red), for two extreme values of $\varepsilon$ (\ie target--source discrepancy). In data-scarce regimes (low $n_T$), our method substantially reduces error (especially the oracle), matching mixed-task behavior when the discrepancy is small. The \citet{ChenOwenShi2014, obst2021transfer, zhang2026unified} struggle to have meaningful improvement compared to task-only. In contrast, our approach never underperforms the target-only baseline (it follows the target-only curve when the discrepancy is large) and yields larger gains when inter-task discrepancy is small.

\tightparagraph{Effect of the noise variance $\sigma_S$:} We inspect the effect of the source observation noise variance $\sigma_S$. We fix the number of target samples and the number of available source samples to the default. We vary the source variance of the source observation from $10^{-3}$ to $10^{3}$, $\varepsilon = 0.2$ and $\sigma_T = 1$. We report the error $\mathrm{err}(\cdot)$ on the left axis and the number of samples borrowed $\wh{n}$ on the right axis.
\begin{SCfigure}[1.1][t]
  \centering
  \includegraphics[width=0.44\linewidth]{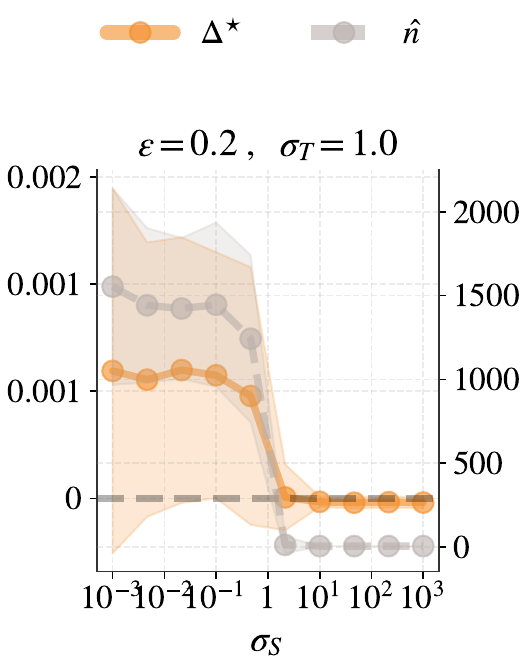}
  \caption{Predictive error comparison (left axis) and the number of samples borrowed $\wh{n}$ (right axis) \wrt the source observation noise variance. The solid line reports the average; while the standard deviation is encoded in transparency.}
  \label{fig:02_sigma_influence__eps_0.2.pdf}
\end{SCfigure}
\autoref{fig:02_sigma_influence__eps_0.2.pdf} plots the error (left axis) and the borrowed source samples $\wh{n}$ (right axis) as a function of the source noise $\sigma_S$. Our method delivers a positive transfer when $\sigma_S \lesssim \sigma_T$, with gains fading as $\sigma_S$ increases; around $\sigma_S \approx \sigma_T$ the algorithm stops borrowing samples ($\wh{n}=0$). This supports the intuition that sharing is beneficial when the source noise is not too dominant.

\tightparagraph{Effect of model discrepancy $\varepsilon = \|\bm{\theta}_S^\star - \bm{\theta}_T^\star\|_2$:}  Additionally, we examine the effect of the difference between the source and target parameters. We fix the number of target samples and the number of available source samples to the default along with the target and source noise variance set to \(\sigma_T=\sigma_S=1\). We vary $\varepsilon$ from $10^{-3}$ to $10^{3}$. We report the error $\mathrm{err}(\cdot)$ on the left axis and the number of samples borrowed $\wh{n}$ on the right axis.
\begin{SCfigure}[1.1][t]
  \centering
  \includegraphics[width=0.44\linewidth]{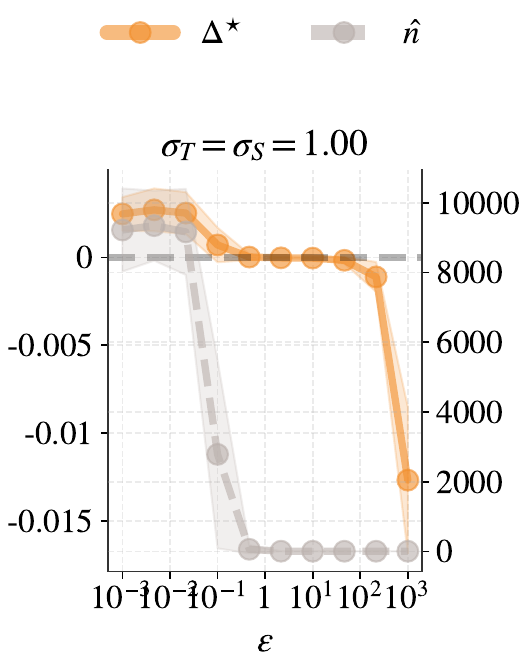}
  \caption{Predictive error comparison (left axis) and the number of samples borrowed $\wh{n}$ (right axis) \wrt the model difference $\varepsilon$. The solid line reports the average; while the standard deviation is encoded in transparency.}
  \label{fig:03_epsilon_influence__sigma_1.00.pdf}
\end{SCfigure}
\autoref{fig:03_epsilon_influence__sigma_1.00.pdf} plots the error (left axis) and the borrowed source samples $\wh{n}$ (right axis) as a function of the distance between tasks $\varepsilon$. Our method yields positive transfer up to $\varepsilon \approx 0.5$, beyond which $\wh{n}$ drops to zero. This underlines the intuitive requirement that the source and target models be sufficiently close for sharing to be beneficial.

\subsection{Real data benchmarks}
\label{subsec:real_data_bench}

We evaluate our method on two real-world regression datasets (Email, and Boston) against established baselines. For each data set, we partition the samples into a \emph{target} task and a \emph{source} task via a clustering-based split (see Appendix~\autoref{sec:supp_exp}), yielding related but non-identical tasks that reflect plausible business scenarios (customers partition). We vary the number of target samples \(n_T\) from \( 100 \) to \( 10000 \), while fixing the pool of source samples at \(n_{\max}=10000\) (so \(\wh{n} \in \{0,\dots,10000\}\)). In each run, we shuffle the training samples for both the target and the source datasets. Each experiment is repeated \(250\) times, and we report averages over runs. As in \autoref{subsec:simulation}, we compare against (i) target-only ridge, (ii) mixed-task ridge, (iii) the data-enriched method of \citet{ChenOwenShi2014}, (iv) the approach of \citet{obst2021transfer},  (v) the approach of \citet{zhang2026unified} and (vi) a hierarchical Bayesian ridge regression baseline. We complete the description of the data set in Appendix~\autoref{sec:supp_exp}.
\begin{figure}[t]
  \centering
  \includegraphics[width=.25\linewidth, trim=0.5cm 11.6cm 6.5cm 0.1cm, clip]{figures/04_performance_error_comparison__eps_0.2.pdf}  
  \includegraphics[width=.32\linewidth, trim=5.5cm 11.6cm 0.5cm 0.9cm, clip]{figures/04_performance_error_comparison__eps_0.2.pdf}  
  \begin{subfigure}{.48\columnwidth}
    \centering
    \includegraphics[width=0.87\linewidth, trim=0.4cm 0.2cm 0.6cm 3.3cm, clip]{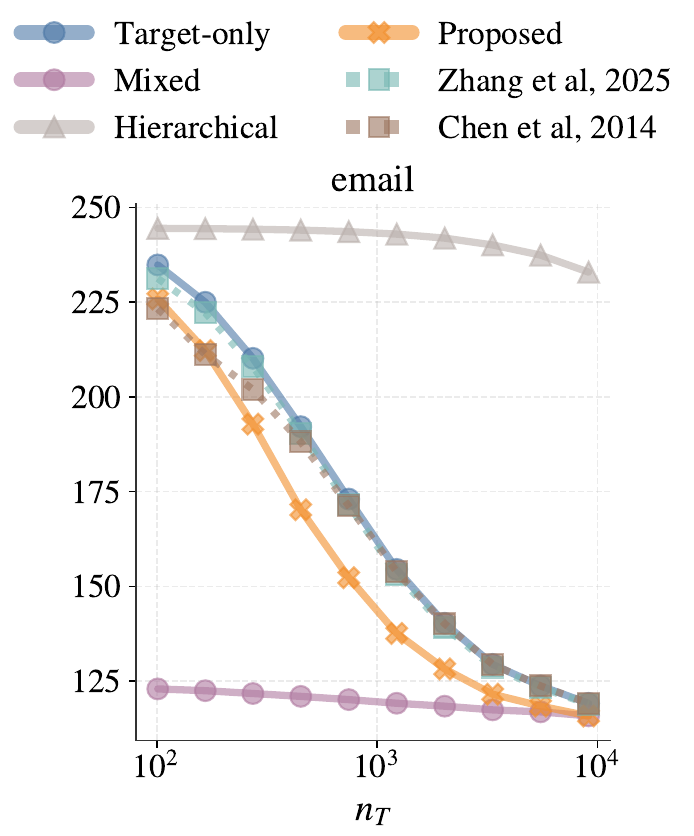}  
    \subcaption{Dataset Email}\label{fig:exp_perf_email}
  \end{subfigure}\hfill%
  \begin{subfigure}{.48\columnwidth}
    \centering
    \includegraphics[width=0.87\linewidth, trim=0.4cm 0.2cm 0.6cm 3.3cm, clip]{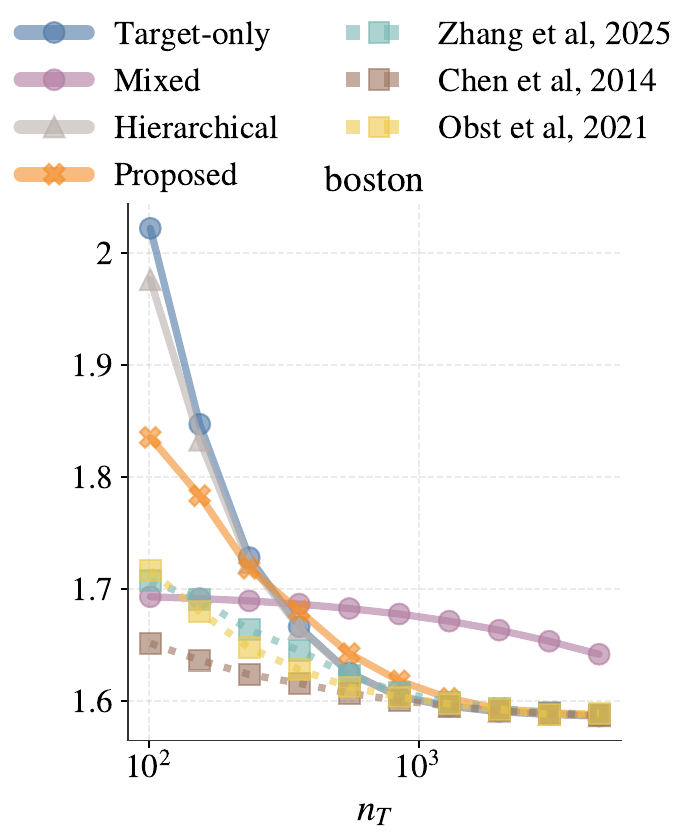}  
    \subcaption{Dataset Boston}\label{fig:exp_perf_boston}
  \end{subfigure}
  \caption{Test-MSE comparison \wrt the number of target samples. The line reports the average.}
  \label{fig:real_data_perf_exp}
\end{figure}
\autoref{fig:real_data_perf_exp} reports the test MSE for the same baselines and color code (except Oracle). Most approaches perform similarly, especially on the Email dataset; for the Boston dataset, they typically avoid the underperformance of Mixed but remain close to target-only. Overall, this suggests that when the linear assumption is misspecified, these methods yield limited improvements, but their main benefit is to prevent large degradations.

We provide additional regression experiments and the classification cases in the Appendix~\autoref{subsec:additional_results}.

\section{Discussion}
\label{sec:discussion}

We further analyze the behavior of the regression transfer gain in this section, along with a broader discussion.\footnote{We focus on regression since it yields better performance; see Appendix~\autoref{subsec:additional_results}.}

\tightparagraph{Asymptotical analysis}
To study the transfer gain, we adopt an isotropic Gaussian design: \( \bm{x}_{T,i} \overset{\iid}{\sim} \mathcal{N}(\bm{0},\I_d)\) and \(\bm{x}_{S,i} \overset{\iid}{\sim} \mathcal{N}(\bm{0},\I_d)\), with features independent across data sets (source \vs target) and independent of the noise.
We consider a source size \(n\to\infty\) with \(d=o(n)\); fixed target sizes \(n_T,n_T^{\mathrm{val}}\); and ridge parameters \(\lambda_T,\lambda_c=O(1)\).
Expectations are taken \wrt the random design matrices \(\bm{X}_T\), \(\bm{X}_T^{\mathrm{val}}\), and \(\bm{X}_S\).

\begin{restatable}[Transfer gain in the large-source isotropic Gaussian regime]{remark}{rma}
    \label{rm:beneficial-collab-large-n-ridge}
    Let \(\bm{\Delta\theta}^\star:=\bm{\theta}_S^\star-\bm{\theta}_T^\star\) and \(\varepsilon^2:=\|\bm{\Delta\theta}^\star\|_2^2\).
    By Marchenko--Pastur concentration, when \(n\to\infty\) (with \(d=o(n)\)), we have:
    {
    \small
    \begin{align*}
         & \mathbb{E}\!\left[\Delta_{\mathrm{regr}}^\star(n)\right]
        \underset{n\to\infty}{=} n_T^{\mathrm{val}} \lambda_T^2 \frac{\|\bm{\theta}_T^\star\|_2^2}{(n_T + \lambda_T)^2}
        + \sigma_T^2 \, n_T^{\mathrm{val}} \frac{d\, n_T}{(n_T + \lambda_T)^2}                                                      \\
         & - n_T^{\mathrm{val}} \frac{\|n\, \bm{\Delta\theta}^\star - \lambda_c \bm{\theta}_T^\star\|_2^2}{(n_T + n + \lambda_c)^2}
        - n_T^{\mathrm{val}} \frac{d\, (\sigma_S^2 n + \sigma_T^2 n_T)}{(n_T + n + \lambda_c)^2} + o(1).
    \end{align*}
    }
\end{restatable}

\tightparagraph{Parameters influence}
Building on~\autoref{rm:beneficial-collab-large-n-ridge}, we quantify how key parameters shape the transfer gain in the regime \(n \gg d\), fixed \(n_T\), and \(\lambda_c=O(1)\), we have:

\begin{enumerate}[leftmargin=2pt,itemsep=1pt,topsep=1pt,parsep=0pt,partopsep=0pt]
    \item \emph{Effect of \(\bm{\Delta\theta}^\star\)}: we have,
          {\small
                  \begin{equation*}
                      \nabla_{\bm{\Delta\theta}^\star} \mathbb{E}\!\left[\Delta_{\mathrm{regr}}^\star(n)\right] = -\frac{n_T^{\mathrm{val}} n}{(n_T+n+\lambda_c)^2}\Big(n \bm{\Delta\theta}^\star-\lambda_c\bm{\theta}_T^\star\Big) + o(1);
                  \end{equation*}
              }
          hence, larger task gaps reduce the gain.
          As $n\to\infty$, the slope approaches $-\;n_T^{\mathrm{val}}$, and it varies linearly with the model mismatch.
    \item \emph{Effect of \(\sigma_T^2\) and \(\sigma_S^2\)}: we have,
          {
                  \small
                  \begin{align*}
                      \frac{\partial}{\partial \sigma_T^2} \mathbb{E}\!\left[\Delta_{\mathrm{regr}}^\star(n)\right] & = n_T^{\mathrm{val}}  d n_T \frac{1}{(n_T+\lambda_T)^2} + o(1),     \\
                      \frac{\partial}{\partial \sigma_S^2} \mathbb{E}\!\left[\Delta_{\mathrm{regr}}^\star(n)\right] & = - n_T^{\mathrm{val}}  d \frac{n}{(n_T + n + \lambda_c)^2} + o(1);
                  \end{align*}
              }
          hence, a higher target noise increases the gain; as \(n\to\infty\), the slope tends to \(n_T^{\mathrm{val}} \; d \; n_T / (n_T + \lambda_T)^2\).
          And, the noisier sources reduce the gain; the marginal harm decays like \(O(1/n)\).
\end{enumerate}

Overall, the most favorable regime features high target noise, strong task alignment (\ie small \(\varepsilon\)), and low source noise.
We now validate these insights with synthetic experiments.

\tightparagraph{General remarks}
We now elaborate on key modeling choices, design rationale, motivating our design, and positioning it against related approaches.

\begin{enumerate}[leftmargin=2pt,itemsep=1pt,topsep=1pt,parsep=0pt,partopsep=0pt]

    \item \emph{Comparison with previous frameworks}: \quad
          Interestingly, our approach can be reformulated to match the~\citet{ChenOwenShi2014, obst2021transfer} framework.
          In fact, both three approaches can be formulated as \( \wh{\bm{\theta}}(n) = \bm{W} \wh{\bm{\theta}}_S(n) + \big( \I_d - \bm{W} \big) \wh{\bm{\theta}}_T\): (a)~\citet{ChenOwenShi2014}: \( \bm{W}(\lambda) = {\bm{\Gamma}(\lambda)}^{-1} \bm{\Psi}(\lambda) \) with \( \bm{\Psi} := \bm{G}_T + \lambda \bm{G}_T^{\mathrm{val}} \bm{G}_S^{-1} \bm{G}_T \;, \bm{\Gamma} := \bm{\Psi} + \lambda \bm{G}_T^{\mathrm{val}} \) and \( \bm{G}_T^{\mathrm{val}} \) the validation Gram matrix.
          (b)~\citet{obst2021transfer}: \( \bm{W}(\alpha, k) = \big( \I_d - \alpha \bm{\Lambda} \big)^k \) with \( \alpha, k \in \mathbb{R}^{+*}\times\mathbb{N}^{*} \) and \( \bm{\Lambda} \) the eigenvalues of \( \bm{G}_T \) the target Gram matrix.
          (c) Our approach: \( \bm{W}(n) = \bm{A}_c(n)^{-1} \bm{A}_S(n) \) and setting \(\lambda_c = \lambda_S + \lambda_T\).
          Hence, all methods introduce \emph{transfer parameters} that control how much information flows from source to target: \( \lambda > 0 \) for~\citet{ChenOwenShi2014}, \( (\alpha>0, \; k\in\mathbb{N}) \) for~\citet{obst2021transfer}, and \(n \in \mathbb{N}\) for ours.
          This parameter is chosen by a data-driven criterion to improve generalization.
          Moreover, only~\citet{obst2021transfer} and our method adopt a conservative policy that transfers only when the estimated gain is positive.
          By contrast, \citet{obst2021transfer} implement transfer via gradient-descent fine-tuning initialized at the source model, offering less transparent control over transfer strength than our sample-sharing mechanism.


    \item \emph{Relying on a validation data set}: \quad
          Our method does require a small validation split to calibrate the decision rule (typically \(\approx 50\) target samples in our experiments), but this cost is modest relative to the benefit.
          In practice, the validation budget pays for itself through consistent error reductions.

\end{enumerate}

\section{Conclusion}
\label{sec:conclusion}

We study dynamic positive transfer for linear prediction with multiple sources. We introduce a target-focused transfer-gain statistic and an online multi-source chunk-selection rule that controls negative transfer using a small validation split. Across synthetic and real datasets, the method improves upon target-only and competitive baselines while selecting when, how much, and from which source to borrow.

\clearpage
\newpage

\begin{acknowledgements}
    This work was supported by the French National Research Agency (ANR) under grant ANR-24-CE40-3341 (project DECATTLON). This work was granted access to the HPC resources of IDRIS under the allocation 2025-AD011016818, provided by GENCI.
\end{acknowledgements}

\bibliographystyle{plainnat}
\renewcommand{\bibsection}{\subsubsection*{References}}
\bibliography{bibliography}

\newpage
\onecolumn

\title{When to Transfer: Adaptive Source Selection for Positive Transfer in Linear Models\\(Supplementary Material)}
\maketitle

\section{Appendix: table of contents}
 \label{sec:supp_table_of_contents}

\begin{enumerate}[align=left,itemsep=0.5cm,label=\Alph*]
    \item Appendix: Notations summary \dotfill \pageref{sec:supp_notations}
    \item Appendix: Estimation errors derivation \dotfill \pageref{sec:supp_est_errors}
    \item Appendix: Characterization of the transfer gain estimator \dotfill \pageref{sec:supp_prop_delta_hat}
    \item Appendix: Lower bound of the transfer gain estimator \dotfill \pageref{sec:supp_lower_bound}
    \item Appendix: Theoretical analysis \dotfill \pageref{sec:supp_theory}
    \item Appendix: Experiments \dotfill \pageref{sec:supp_exp}
\end{enumerate}

\section{Appendix: Notations summary}
\label{sec:supp_notations}

\renewcommand{\arraystretch}{1.5}

\begin{longtable}{@{}p{0.23\textwidth} p{0.73\textwidth}@{}}
    \textbf{Symbol}                                                                                                                 & \textbf{Description}                                                                                                                \\
    \hline
    \endfirsthead
    \textbf{Symbol}                                                                                                                 & \textbf{Description}                                                                                                                \\
    \hline
    \endhead
    $n_T$, $n_T^{\mathrm{val}}$,$n_T^{\mathrm{test}}$                                                                               & Number of target \emph{training}, \emph{validation} and \emph{test} samples.                                                        \\
    \hline
    $n_{\max}$                                                                                                                      & Maximum number of source samples allowed to be borrowed (budget).                                                                   \\
    \hline
    $n$                                                                                                                             & Number of source samples used in sharing.                                                                                           \\
    \hline
    $\delta$, $\alpha$                                                                                                              & Confidence level, practical borrowing parameter.                                                                                    \\
    \hline
    $\lambda_T,\ \lambda_S,\ \lambda_c$                                                                                             & Ridge regularization parameters (target / source / collaborative).                                                                  \\
    \hline
    $\sigma_T^2,\ \sigma_S^2$                                                                                                       & Noise variances on target/source.                                                                                                   \\
    \hline
    $\bm{\theta}_T^\star,\ \bm{\theta}_S^\star\in\mathbb{R}^d$                                                                      & Unknown target/source ground-truth parameters.                                                                                      \\
    \hline
    $\Delta\bm{\theta}^\star := \bm{\theta}_S^\star-\bm{\theta}_T^\star$, $\varepsilon := \|\Delta\bm{\theta}^\star\|_2$            & Inter-task parameter shift and parameter distance.                                                                                  \\
    \hline
    $\bm{\eta}_T\in\mathbb{R}^{n_T},\ \bm{\eta}_S\in\mathbb{R}^{n}$                                                                 & Target/source noise vectors (mean $0$, covariances $\sigma_T^2\I,\sigma_S^2\I$).                                                    \\
    \hline
    $\bm{X}_T\in\mathbb{R}^{n_T\times d},\ \bm{y}_T\in\mathbb{R}^{n_T}$                                                             & Target training design matrix and responses (or labels for classification).                                                         \\
    \hline
    $\bm{X}_T^{\mathrm{val}}\in\mathbb{R}^{n_T^{\mathrm{val}}\times d},\ \bm{y}_T^{\mathrm{val}}\in\mathbb{R}^{n_T^{\mathrm{val}}}$ & Target validation set (features observed; used for criteria).                                                                       \\
    \hline
    $\bm{X}_T^{\mathrm{test}}\in\mathbb{R}^{n_T^{\mathrm{test}}\times d}$                                                           & Target test design matrix (used for empirical test risk).                                                                           \\
    \hline
    $\bm{X}_S(n)\in\mathbb{R}^{n\times d},\ \bm{y}_S(n)\in\mathbb{R}^{n}$                                                           & Matrix/vector formed by the first $n$ source samples and responses.                                                                 \\
    \hline
    $\bm{X}_c\in\mathbb{R}^{(n_T+n)\times d},\ \bm{y}_c\in\mathbb{R}^{n_T+n}$                                                       & Stacked (target+source) design matrix and responses.                                                                                \\
    \hline
    $\bm{G}_T := \bm{X}_T^\top \bm{X}_T$                                                                                            & Target (unregularized) Gram/design matrix.                                                                                          \\
    \hline
    $\bm{G}_S(n) := \sum_{j=1}^{n}\bm{x}_{S,j}\bm{x}_{S,j}^\top$                                                                    & Source Gram built from the first $n$ samples.                                                                                       \\
    \hline
    $\bm{b}_T := \bm{X}_T^\top \bm{y}_T$                                                                                            & Target cross term.                                                                                                                  \\
    \hline
    $\bm{b}_S(n) := \sum_{j=1}^{n}\bm{x}_{S,j}\,y_{S,j}$                                                                            & Source cross term from the first $n$ samples.                                                                                       \\
    \hline
    $\bm{Z}_T := \bm{X}_T^\top \bm{\eta}_T$                                                                                         & Target noise cross term.                                                                                                            \\
    \hline
    $\bm{Z}_S(n) := \bm{X}_S(n)^\top \bm{\eta}_S(n)$                                                                                & Source noise cross term (first $n$ noises).                                                                                         \\
    \hline
    $\bm{A}_T := \bm{G}_T + \lambda_T\I_d$                                                                                          & Target regularized Gram; $\bm{A}_T^{-1}$ its inverse.                                                                               \\
    \hline
    $\bm{A}_S(n)$                                                                                                                   & Source regularized Gram; $\bm{A}_S(n)^{-1}$ its inverse.                                                                            \\
    \hline
    $\bm{A}_c(n) := \bm{G}_T + \bm{G}_S(n) + \lambda_{c}\I_d$                                                                       & Joint/collaborative regularized Gram; $\bm{A}_c(n)^{-1}$ its inverse.                                                               \\
    \hline
    $\wh{\bm{\theta}}_{T} := \bm{A}_T^{-1}\bm{b}_T$, $\wh{\bm{\theta}}_S(n)$                                                        & Target (\resp source) ridge estimator.                                                                                              \\
    \hline
    $\wh{\bm{\theta}}(n) := \bm{A}_c(n)^{-1}\big(\bm{b}_T+\bm{b}_S(n)\big)$                                                         & Collaborative ridge estimator using $n$ source samples.                                                                             \\
    \hline
    $\bm{U}_T := \bm{X}_T^{\mathrm{val}}\bm{A}_T^{-1}$                                                                              & Validation-to-parameter linear map for target estimator.                                                                            \\
    \hline
    $\bm{V}(n) := \bm{X}_T^{\mathrm{val}}\bm{A}_c(n)^{-1}$                                                                          & Validation-to-parameter linear map for collaborative estimator.                                                                     \\
    \hline
    $\xi_T$, $\zeta_T$                                                                                                              & Target-only \emph{regression} (and \emph{classification}) error on validation set.                                                  \\
    \hline
    $\xi(n)$ , $\zeta(n)$                                                                                                           & Collaborative \emph{regression} (and \emph{classification})) error on validation set.                                               \\
    \hline
    $\Delta^\star_{\mathrm{regr}}(n):=\xi_T-\xi(n)$, $\Delta^\star_{\mathrm{clf}}(n):=\zeta_T-\zeta(n)$                             & True regression (\resp classification) transfer gain.                                                                               \\
    \hline
    $\wh{\Delta}_{\mathrm{regr}}(n)$, $\wh{\Delta}_{\mathrm{clf}}(n)$                                                               & Plug-in estimator of $\Delta^\star_{\mathrm{regr}}(n)$ (\resp $\Delta^\star_{\mathrm{clf}}(n)$).                                    \\
    \hline
    $\mathrm{LCB}^{\mathrm{regr}}_\delta(n)$, $\mathrm{LCB}^{\mathrm{clf}}_\delta(n)$                                               & One-sided lower confidence bound for $\Delta^\star_{\mathrm{regr}}(n)$ (\resp $\Delta^\star_{\mathrm{clf}}(n)$).                    \\
    \hline
    $\mathrm{TRGA}_{\alpha,n}$                                                                                                      & Practical decision statistic: $\wh{\Delta}_{\mathrm{regr}}(n)-\alpha\sqrt{\widehat{\mathrm{Var}}(\wh{\Delta}_{\mathrm{regr}}(n))}$. \\
    \hline
    $\Phi(\cdot)$                                                                                                                   & Probit surrogate (standard normal CDF), applied elementwise.                                                                        \\
    \hline
    $\odot,\ \oslash$                                                                                                               & Hadamard (elementwise) product and division.                                                                                        \\
    \hline
    $g_T := \bm{G}_T\bm{\theta}_T^\star$                                                                                            & Target signal term in classification formulas.                                                                                      \\
    \hline
    $g_c(n):=\bm{G}_T\bm{\theta}_T^\star+\bm{G}_S(n)\bm{\theta}_S^\star$                                                            & Collaborative signal term in classification formulas.                                                                               \\
    \hline
    $s_T^2 := \sigma_T^2\,\mathrm{diag}(\bm{U}_T\bm{G}_T\bm{U}_T^\top)$                                                             & Per-validation-point predictive variance (target-only).                                                                             \\
    \hline
    $s_T^2(n) := \sigma_T^2\,\mathrm{diag}(\bm{V}(n)\bm{G}_T\bm{V}(n)^\top)$                                                        & Target-noise contribution to collaborative predictive variance.                                                                     \\
    \hline
    $s_S^2(n) := \sigma_S^2\,\mathrm{diag}(\bm{V}(n)\bm{G}_S(n)\bm{V}(n)^\top)$                                                     & Source-noise contribution to collaborative predictive variance.                                                                     \\
    \hline
    $s_c^2(n):=s_T^2(n)+s_S^2(n)$                                                                                                   & Total collaborative predictive variance (classification).                                                                           \\
    \hline
    $S,\ s$                                                                                                                         & Number of sources and source index (multi-source setting).                                                                          \\
\end{longtable}

\section{Appendix: Prediction errors derivation}
\label{sec:supp_est_errors}

We provide explicit derivations of the terms for single-task and collaborative error.

\subsection{Regression task}

\paragraph{Single error:} The single task prediction error \( \xi_T := \mathbb{E}\!\left[ \left\| \bm{X}_T^{\mathrm{test}} \big( \wh{\bm{\theta}}_{T} \;-\; \bm{\theta}_T^\star \big) \right\|_2^2 \right] \) can be expanded as:

\begin{align*}
    \xi_T
     & := \mathbb{E}\!\left[ \left\| \bm{X}_T^{\mathrm{test}} \big( \wh{\bm{\theta}}_{T} \;-\; \bm{\theta}_T^\star \big) \right\|_2^2 \right]                                                                                     \\
     & = \mathbb{E}\!\left[ \left\| \bm{X}_T^{\mathrm{test}} \big( \big( {\bm{A}_T}^{-1} \bm{G}_T \bm{\theta}_T^\star + {\bm{A}_T}^{-1} \bm{Z}_T \big) \;-\; \bm{\theta}_T^\star \big) \right\|_2^2 \right]                       \\
     & = \mathbb{E}\!\left[ \left\| \bm{X}_T^{\mathrm{test}} \big( {\bm{A}_T}^{-1} \bm{G}_T - \I_d \big) \bm{\theta}_T^\star + \bm{X}_T^{\mathrm{test}} {\bm{A}_T}^{-1} \bm{Z}_T \right\|_2^2 \right]                             \\
     & = \mathbb{E}\!\left[ \left\| \bm{X}_T^{\mathrm{test}}  {\bm{A}_T}^{-1} \big( \bm{G}_T - {\bm{A}_T} \big) \bm{\theta}_T^\star + \bm{X}_T^{\mathrm{test}} {\bm{A}_T}^{-1} \bm{Z}_T \right\|_2^2 \right]                      \\
     & = \mathbb{E}\!\left[ \left\| - \lambda\bm{X}_T^{\mathrm{test}}  {\bm{A}_T}^{-1} \bm{\theta}_T^\star + \bm{X}_T^{\mathrm{test}} {\bm{A}_T}^{-1} \bm{Z}_T \right\|_2^2 \right]                                               \\
     & = \left\| - \lambda\bm{X}_T^{\mathrm{test}}  {\bm{A}_T}^{-1} \bm{\theta}_T^\star  \right\|_2^2 +  \mathbb{E}\!\left[ \left\| \bm{X}_T^{\mathrm{test}} {\bm{A}_T}^{-1} \bm{Z}_T \right\|_2^2 \right]                        \\
     & = \lambda^2\left\| \bm{X}_T^{\mathrm{test}}{\bm{A}_T}^{-1}\bm{\theta}_T^\star \right\|_2^2 \;+\;\sigma_T^2\,\tr\!\big(\bm{X}_T^{\mathrm{test}}{\bm{A}_T}^{-1}\bm{G}_T{\bm{A}_T}^{-1}{\bm{X}_T^{\mathrm{test}}}^\top\big)\,
\end{align*}

\paragraph{Collaboration error:} the collaborative prediction error \( \xi(n) := \mathbb{E}\! \left[ \left\|  \bm{X}_T^{\mathrm{test}} (\wh{\bm{\theta}}(n) - \bm{\theta}_T^\star \big) \right\|_2^2 \right] \) can be expanded as:

\begin{align*}
    \xi(n)
     & := \mathbb{E}\! \left[ \left\|  \bm{X}_T^{\mathrm{test}} (\wh{\bm{\theta}}(n) - \bm{\theta}_T^\star \big) \right\|_2^2 \right]                                                                                                                                                                                                                                                                     \\
     & = \mathbb{E}\!\left[ \left\| \bm{X}_T^{\mathrm{test}} \big( {\bm{A_c}(n)}^{-1} \big( \bm{G}_T \bm{\theta}_T^\star + \bm{G}_S(n) \bm{\theta}_S^\star + \bm{Z}_S(n) + \bm{Z}_T \big) - \bm{\theta}_T^\star \big) \right\|_2^2 \right]                                                                                                                                                                \\
     & = \mathbb{E}\!\left[ \left\| \bm{X}_T^{\mathrm{test}} {\bm{A}_c(n)}^{-1} \big( \bm{G}_T \bm{\theta}_T^\star + \bm{G}_S(n) \bm{\theta}_S^\star - \bm{A}_c(n) \bm{\theta}_T^\star \big) + \bm{X}_T^{\mathrm{test}} {\bm{A}_c(n)}^{-1} \big( \bm{Z}_S(n) + \bm{Z}_T \big)  \right\|_2^2 \right]                                                                                                       \\
     & = \left\| \bm{X}_T^{\mathrm{test}} {\bm{A}_c(n)}^{-1} \big( \bm{G}_T \bm{\theta}_T^\star + \bm{G}_S(n) \bm{\theta}_S^\star - \bm{A}_c(n) \bm{\theta}_T^\star \big)  \right\|_2^2  + \mathbb{E}\!\left[ \left\| \bm{X}_T^{\mathrm{test}} {\bm{A}_c(n)}^{-1} \bm{Z}_S(n) \right\|_2^2 \right] + \mathbb{E}\!\left[ \left\| \bm{X}_T^{\mathrm{test}} {\bm{A}_c(n)}^{-1} \bm{Z}_T \right\|_2^2 \right] \\
     & = \left\| \bm{X}_T^{\mathrm{test}} {\bm{A}_c(n)}^{-1} \big( \bm{G}_S(n) \big( \bm{\theta}_S^\star - \bm{\theta}_T^\star \big)- \lambda \bm{\theta}_T^\star \big)  \right\|_2^2  + \mathbb{E}\!\left[ \left\| \bm{X}_T^{\mathrm{test}} {\bm{A}_c(n)}^{-1} \bm{Z}_S(n) \right\|_2^2 \right] + \mathbb{E}\!\left[ \left\| \bm{X}_T^{\mathrm{test}} {\bm{A}_c(n)}^{-1} \bm{Z}_T \right\|_2^2 \right]   \\
     & = \left\| \bm{X}_T^{\mathrm{test}} \bm{A}_c(n)^{-1}\!\left(\bm{G}_S(n)\,(\bm{\theta}_S^\star - \bm{\theta}_T^\star)-\lambda\,\bm{\theta}_T^\star\right) \right\|_2^2 + \tr\!\Big(\bm{X}_T^{\mathrm{test}} \bm{A}_c(n)^{-1}\big(\sigma_S^2 \bm{G}_S(n)+\sigma_T^2 \bm{G}_T\big)\bm{A}_c(n)^{-1}{\bm{X}_T^{\mathrm{test}}}^\top \Big).
\end{align*}

\tightparagraph{Target-only classification}
Assume that $\bm{\pi}:=\mathbb{P}(\bm{y}_T^{\mathrm{val}}=\bm{1}\mid \bm{X}_T^{\mathrm{val}})$.
    {\small
        \begin{align*}
            \zeta_T
            :=\, & \frac{1}{n_T^{\mathrm{val}}}\, \mathbb{E}\!\left[\bm{1}^\top \Phi\!\Big( -\,\bm{y}_T^{\mathrm{val}}\odot(\bm{X}_T^{\mathrm{val}}\wh{\bm{\theta}}_T) \Big) \right]                                          \\
            =\,  & \frac{1}{n_T^{\mathrm{val}}}\, \mathbb{E}\!\left[\bm{1}^\top \Phi\!\Big( -\,\bm{y}_T^{\mathrm{val}}\odot\big(\bm{X}_T^{\mathrm{val}}\bm{A}_T^{-1}(\bm{G}_T\bm{\theta}_T^\star+\bm{Z}_T)\big) \Big) \right] \\
            =\,  & \frac{1}{n_T^{\mathrm{val}}}\, \mathbb{E}\!\left[\bm{1}^\top \Phi\!\Big( -\,\bm{y}_T^{\mathrm{val}}\odot\big(\bm{U}_T\bm{g}_T+\bm{U}_T\bm{Z}_T\big) \Big) \right]                                          \\
            =\,  & \frac{1}{n_T^{\mathrm{val}}}\, \mathbb{E}\!\left[\bm{1}^\top \Phi\!\left( \Big(-\,\bm{y}_T^{\mathrm{val}}\odot(\bm{U}_T\bm{g}_T)\Big)\oslash\sqrt{\bm{1}+\bm{s}_T^2} \right) \right]                       \\
            =\,  & \frac{1}{n_T^{\mathrm{val}}}\, \bm{1}^\top\Big( \bm{\pi}\odot \Phi(-\bm{t}_T)\;+\;(\bm{1}-\bm{\pi})\odot \Phi(\bm{t}_T) \Big),
        \end{align*}
    }
with $\bm{U}_T:=\bm{X}_T^{\mathrm{val}}\bm{A}_T^{-1}$, $\bm{g}_T:=\bm{G}_T\bm{\theta}_T^\star$,
$\bm{Z}_T:=\bm{X}_T^\top\bm{\eta}_T$, $\bm{s}_T^2:=\sigma_T^2\,\diag(\bm{U}_T\bm{G}_T\bm{U}_T^\top)$, and
$\bm{t}_T:= (\bm{U}_T\bm{g}_T)\oslash\sqrt{\bm{1}+\bm{s}_T^2}$.

\tightparagraph{Collaborative classification}
Assume that $\bm{\pi}:=\mathbb{P}(\bm{y}_T^{\mathrm{val}}=\bm{1}\mid \bm{X}_T^{\mathrm{val}})$.
    {\small
        \begin{align*}
            \zeta(n)
            :=\, & \frac{1}{n_T^{\mathrm{val}}}\, \mathbb{E}\!\left[\bm{1}^\top \Phi\!\Big( -\,\bm{y}_T^{\mathrm{val}}\odot(\bm{X}_T^{\mathrm{val}}\wh{\bm{\theta}}(n)) \Big) \right]                                        \\
            =\,  & \frac{1}{n_T^{\mathrm{val}}}\, \mathbb{E}\!\left[\bm{1}^\top \Phi\!\Big( -\,\bm{y}_T^{\mathrm{val}}\odot\big(\bm{X}_T^{\mathrm{val}}\bm{A}_c(n)^{-1}(\bm{g}_c(n)+\bm{Z}_T+\bm{Z}_S(n))\big) \Big) \right] \\
            =\,  & \frac{1}{n_T^{\mathrm{val}}}\, \mathbb{E}\!\left[\bm{1}^\top \Phi\!\Big( -\,\bm{y}_T^{\mathrm{val}}\odot\big(\bm{V}(n)\bm{g}_c(n)+\bm{V}(n)(\bm{Z}_T+\bm{Z}_S(n))\big) \Big) \right]                      \\
            =\,  & \frac{1}{n_T^{\mathrm{val}}}\, \mathbb{E}\!\left[\bm{1}^\top \Phi\!\left( \Big(-\,\bm{y}_T^{\mathrm{val}}\odot(\bm{V}(n)\bm{g}_c(n))\Big)\oslash\sqrt{\bm{1}+\bm{s}_c^2(n)} \right) \right]               \\
            =\,  & \frac{1}{n_T^{\mathrm{val}}}\, \bm{1}^\top\Big( \bm{\pi}\odot \Phi(-\bm{t}_c(n))\;+\;(\bm{1}-\bm{\pi})\odot \Phi(\bm{t}_c(n)) \Big),
        \end{align*}
    }
where $\bm{V}(n):=\bm{X}_T^{\mathrm{val}}\bm{A}_c(n)^{-1}$,
$\bm{g}_c(n):=\bm{G}_T\bm{\theta}_T^\star+\bm{G}_S(n)\bm{\theta}_S^\star$,
$\bm{s}_T^2(n):=\sigma_T^2\,\diag(\bm{V}(n)\bm{G}_T\bm{V}(n)^\top)$,
$\bm{s}_S^2(n):=\sigma_S^2\,\diag(\bm{V}(n)\bm{G}_S(n)\bm{V}(n)^\top)$,
$\bm{s}_c^2(n):=\bm{s}_T^2(n)+\bm{s}_S^2(n)$, and
$\bm{t}_c(n):=(\bm{V}(n)\bm{g}_c(n))\oslash\sqrt{\bm{1}+\bm{s}_c^2(n)}$.

\section{Appendix: Characterization of the transfer gain estimator}
\label{sec:supp_prop_delta_hat}

We prove the characterization of the regression transfer gain estimator:

\propa*

\begin{proof}[Proof of {\upshape\autoref{prop:expect_var_delta_hat_regr}}]

    By expanding the expectation, we have the following.
    \begin{align*}
        \mathbb{E}\!\left[\wh{\Delta}_{\mathrm{regr}}(n)\right]
         & = \lambda_T^2\,\mathbb{E}\!\left[\left\| \bm{U}_T \wh{\bm{\theta}}_T \right\|_2^2\right]
        - \mathbb{E}\!\left[\left\| \bm{V}(n)\!\left( \bm{G}_S(n)( \wh{\bm{\theta}}_S - \wh{\bm{\theta}}_T ) - \lambda_c \wh{\bm{\theta}}_T \right) \right\|_2^2\right]
        + \tr\!\big( \bm{U}_T \bm{M}_T \bm{U}_T^{\top} \big) - \tr\!\big( \bm{V}(n) \bm{N}(n) \bm{V}(n)^{\top} \big)                                                                    \\
         & = \lambda_T^2\,\mathbb{E}\!\left[\left\| \bm{U}_T\!\left(\bm{\theta}_T^\star - \lambda_T \bm{A}_T^{-1}\bm{\theta}_T^\star + \bm{A}_T^{-1}\bm{Z}_T\right) \right\|_2^2\right] \\
         & - \mathbb{E}\!\left[\left\| \bm{V}(n)\!\left(
        \bm{G}_S(n)\!\left(\bm{\Delta\theta}-\bm{\Delta\theta}^{\mathrm b}\right)
        - \lambda_c \bm{\theta}_T^\star + \lambda_c \lambda_T \bm{A}_T^{-1}\bm{\theta}_T^\star
        + \bm{G}_S(n)\bm{A}_S(n)^{-1}\bm{Z}_S(n) - (\bm{G}_S(n)+\lambda_c\I_d)\bm{A}_T^{-1}\bm{Z}_T \right)\right\|_2^2\right]                                                          \\
    \end{align*}

    We further expand the second term.

    \begin{align*}
         & = \lambda_T^2\Big\| \bm{U}_T\!\big(\bm{\theta}_T^\star - \lambda_T \bm{A}_T^{-1}\bm{\theta}_T^\star\big) \Big\|_2^2 + \lambda_T^2\,\tr\!\big(\bm{U}_T\bm{A}_T^{-1}\,\mathbb{E}[\bm{Z}_T\bm{Z}_T^\top]\,\bm{A}_T^{-1}\bm{U}_T^\top\big)                                                                               \\
         & \quad - \Big\| \bm{V}(n)\!\Big( \bm{G}_S(n)\bm{\Delta\theta}-\lambda_c\bm{\theta}_T^\star \;-\; \bm{G}_S(n)\bm{\Delta\theta}^{\mathrm b}-\lambda_c\lambda_T \bm{A}_T^{-1}\bm{\theta}_T^\star\Big) \Big\|_2^2                                                                                                         \\
         & \quad - \tr\!\Big(\bm{V}(n)\Big[ \mathbb{E}\!\big[(\bm{G}_S\bm{A}_S^{-1}\bm{Z}_S)(\bm{G}_S\bm{A}_S^{-1}\bm{Z}_S)^\top\big] + \mathbb{E}\!\big[((\bm{G}_S+\lambda_c\I_d)\bm{A}_T^{-1}\bm{Z}_T)((\bm{G}_S+\lambda_c\I_d)\bm{A}_T^{-1}\bm{Z}_T)^\top\big] \Big]\bm{V}(n)^\top\Big)                                      \\
         & = \lambda_T^2\Big\| \bm{U}_T\!\big(\bm{\theta}_T^\star - \lambda_T \bm{A}_T^{-1}\bm{\theta}_T^\star\big) \Big\|_2^2 + \lambda_T^2\,\tr\!\big(\bm{U}_T\bm{A}_T^{-1}\sigma_T^2\bm{G}_T\bm{A}_T^{-1}\bm{U}_T^\top\big)                                                                                                  \\
         & \quad - \Big\| \bm{V}(n)\!\Big( \bm{G}_S(n)\bm{\Delta\theta}-\lambda_c\bm{\theta}_T^\star - \big(\bm{G}_S(n)\bm{\Delta\theta}^{\mathrm b}-\lambda_c\lambda_T \bm{A}_T^{-1}\bm{\theta}_T^\star\big) \Big) \Big\|_2^2                                                                                                  \\
         & \quad - \tr\!\Big(\bm{V}(n)\Big[ \sigma_S^2\,\bm{G}_S(n)\bm{A}_S(n)^{-1}\bm{G}_S(n)\bm{A}_S(n)^{-1}\bm{G}_S(n) + \sigma_T^2\,\big(\bm{G}_S(n)+\lambda_c\I_d\big)\bm{A}_T^{-1}\bm{G}_T\bm{A}_T^{-1}\big(\bm{G}_S(n)+\lambda_c\I_d\big) \Big]\bm{V}(n)^\top\Big)                                                       \\
         & \quad + \tr\!\big( \bm{U}_T \bm{M}_T \bm{U}_T^{\top} \big) - \tr\!\big( \bm{V}(n) \bm{N}(n) \bm{V}(n)^{\top} \big)                                                                                                                                                                                                   \\
         & = \lambda_T^2\|\bm{U}_T\bm{\theta}_T^\star\|_2^2 - \|\bm{V}(n)(\bm{G}_S(n)\bm{\Delta\theta}-\lambda_c\bm{\theta}_T^\star)\|_2^2 + \lambda_T^4\|\bm{U}_T\bm{A}_T^{-1}\bm{\theta}_T^\star\|_2^2 - 2\lambda_T^2\,\big\langle \bm{U}_T\bm{\theta}_T^\star,\;\lambda_T\bm{U}_T\bm{A}_T^{-1}\bm{\theta}_T^\star\big\rangle \\
         & \quad - \|\bm{V}(n)(\bm{G}_S(n)\bm{\Delta\theta}^{\mathrm b}-\lambda_c\lambda_T\bm{A}_T^{-1}\bm{\theta}_T^\star)\|_2^2                                                                                                                                                                                               \\
         & \quad + 2\,\Big\langle \bm{V}(n)\big(\bm{G}_S(n)\bm{\Delta\theta}-\lambda_c\bm{\theta}_T^\star\big),\; \bm{V}(n)\big(\bm{G}_S(n)\bm{\Delta\theta}^{\mathrm b}-\lambda_c\lambda_T\bm{A}_T^{-1}\bm{\theta}_T^\star\big)\Big\rangle                                                                                     \\
         & \quad + \Big[\lambda_T^2\,\tr\!\big(\bm{U}_T\bm{A}_T^{-1}\sigma_T^2\bm{G}_T\bm{A}_T^{-1}\bm{U}_T^\top\big) + \tr\!\big( \bm{U}_T \bm{M}_T \bm{U}_T^{\top} \big)\Big]                                                                                                                                                 \\
         & \quad - \Big[\tr\!\Big(\bm{V}(n)\big(\sigma_S^2\,\bm{G}_S\bm{A}_S^{-1}\bm{G}_S\bm{A}_S^{-1}\bm{G}_S + \sigma_T^2\,(\bm{G}_S+\lambda_c\I_d)\bm{A}_T^{-1}\bm{G}_T\bm{A}_T^{-1}(\bm{G}_S+\lambda_c\I_d) \big)\bm{V}(n)^\top\Big) + \tr\!\big( \bm{V}(n) \bm{N}(n) \bm{V}(n)^{\top} \big)\Big]                           \\
         & = \lambda_T^2\|\bm{U}_T\bm{\theta}_T^\star\|_2^2 - \|\bm{V}(n)(\bm{G}_S(n)\bm{\Delta\theta}-\lambda_c\bm{\theta}_T^\star)\|_2^2                                                                                                                                                                                      \\
         & \quad + \sigma_T^2\,\tr\!\big(\bm{U}_T\bm{A}_T^{-1}\bm{G}_T\bm{A}_T^{-1}\bm{U}_T^\top\big) - \tr\!\Big(\bm{V}(n)\big(\sigma_S^2 \bm{G}_S(n)+\sigma_T^2 \bm{G}_T\big)\bm{V}(n)^\top\Big)                                                                                                                              \\
         & \quad + \lambda_T^4\|\bm{U}_T\bm{A}_T^{-1}\bm{\theta}_T^\star\|_2^2 - 2\lambda_T^2\,\big\langle \bm{U}_T\bm{\theta}_T^\star,\;\lambda_T\bm{U}_T\bm{A}_T^{-1}\bm{\theta}_T^\star\big\rangle - \|\bm{V}(n)(\bm{G}_S(n)\bm{\Delta\theta}^{\mathrm b}-\lambda_c\lambda_T\bm{A}_T^{-1}\bm{\theta}_T^\star)\|_2^2          \\
         & \quad - \Big\langle \bm{V}(n)\big(\bm{G}_S(n)\bm{\Delta\theta}-\lambda_c\bm{\theta}_T^\star\big),\; \bm{V}(n)\big(\bm{G}_S(n)\bm{\Delta\theta}^{\mathrm b}+\lambda_c\lambda_T\bm{A}_T^{-1}\bm{\theta}_T^\star\big)\Big\rangle                                                                                        \\
    \end{align*}

    Which leads to:
    \begin{equation*}
        \mathbb{E}\!\left[\wh{\Delta}_{\mathrm{regr}}(n)\right] = \Delta_{\mathrm{regr}}^\star(n) \;+\; b(n,\lambda_c,\lambda_S,\lambda_T),
    \end{equation*}

    We now make explicit the closed form of \( \var\big( \wh{\Delta}_{\mathrm{regr}}(n) \big) \). Since the deterministic terms do not affect the variance, we collect them into a constant \(c(n)\). From \autoref{def:estimated_transfert_grain_regr},
    \[
        \wh{\Delta}_{\mathrm{regr}}(n) = \lambda_T^2 \big\| \bm{U}_T \wh{\bm{\theta}}_T \big\|_2^2 \;-\; \big\| \bm{V}(n)\big( \bm{G}_S(n)\big(\wh{\bm{\theta}}_S - \wh{\bm{\theta}}_T\big) - \lambda_c \wh{\bm{\theta}}_T \big) \big\|_2^2 \;+\; c(n), \quad \bm{U}_T := \bm{X}_T^{\mathrm{val}} \bm{A}_T^{-1},\ \ \bm{V}(n) := \bm{X}_T^{\mathrm{val}} \bm{A}_c(n)^{-1}.
    \]
    Expanding yields
    \begin{align*}
        \wh{\Delta}_{\mathrm{regr}}(n)
         & = \wh{\bm{\theta}}_T^\top \Big( \lambda_T^2 \bm{U}_T^\top \bm{U}_T \Big) \wh{\bm{\theta}}_T \;-\; \| \bm{V}(n) \bm{G}_S(n) \wh{\bm{\theta}}_S(n) - \bm{V}(n) \Big( \bm{G}_S(n) + \lambda_c \I_d \Big) \wh{\bm{\theta}}_T \|_2^2  \;+\; c(n) \\
         & = \wh{\bm{\theta}}_T^\top \Big( \lambda_T^2 \bm{U}_T^\top \bm{U}_T \Big) \wh{\bm{\theta}}_T \;-\; \wh{\bm{\theta}}_S(n)^\top \Big( \bm{G}_S(n)^\top \bm{V}(n)^\top \bm{V}(n) \bm{G}_S(n) \Big) \wh{\bm{\theta}}_S                           \\
         & \qquad \;-\; \wh{\bm{\theta}}_T^\top \Big( \bm{G}_S(n) + \lambda_c \I_d \Big)^\top \bm{V}(n)^\top \bm{V}(n) \Big( \bm{G}_S(n) + \lambda_c \I_d \Big) \wh{\bm{\theta}}_T                                                                     \\
         & \qquad \;+\; 2 \wh{\bm{\theta}}_S(n)^\top \Big( \bm{G}_S(n)^\top \bm{V}(n)^\top \bm{V}(n) \Big( \bm{G}_S(n) + \lambda_c \I_d \Big)  \Big) \wh{\bm{\theta}}_T \;+\; c(n)                                                                     \\
    \end{align*}
    Let us set:
    \begin{align*}
        \bm{D}_{11}(n) & = \;-\; \bm{G}_S(n)^\top \bm{V}(n)^\top \bm{V}(n)\,\bm{G}_S(n) \;,                                                                                            \\
        \bm{D}_{12}(n) & = \bm{G}_S(n)^\top \bm{V}(n)^\top \bm{V}(n) \; \big(\bm{G}_S(n) + \lambda_c \I_d\big) \;,                                                                     \\
        \bm{D}_{21}(n) & = \big(\bm{G}_S(n) + \lambda_c \I_d\big)^\top \bm{V}(n)^\top \bm{V}(n) \; \bm{G}_S(n) \;,                                                                     \\
        \bm{D}_{22}(n) & = \lambda_T^2 \; \bm{U}_T^\top \bm{U}_T \;-\; \big(\bm{G}_S(n) + \lambda_c \I_d\big)^\top \bm{V}(n)^\top \bm{V}(n) \; \big(\bm{G}_S(n) + \lambda_c \I_d\big).
    \end{align*}
    This give us:
    \[
        \wh{\Delta}_{\mathrm{regr}}(n)
        \;=\; \wh{\bm{\theta}}_T^\top \bm{D}_{11}(n) \wh{\bm{\theta}}_T
        \;+\; \wh{\bm{\theta}}_S(n)^\top \bm{D}_{22}(n) \wh{\bm{\theta}}_S(n)
        \;+\; \wh{\bm{\theta}}_S(n)^\top \bm{D}_{12}(n)
        \;+\; \bm{D}_{21}(n) \wh{\bm{\theta}}_T
        \;+\; c(n),
    \]
    which leads to \( \wh{\Delta}_{\mathrm{regr}}(n) = \bm{z}(n)^\top \; \bm{D}(n) \; \bm{z}(n) + c(n) \quad\mathrm{with}\quad \bm{z}(n) = \Big[ \wh{\bm{\theta}}_S(n)^\top \; \wh{\bm{\theta}}_T^\top \Big]^\top \) and
    \( \bm{D}(n) = \begin{bmatrix} \bm{D}_{11}(n) & \bm{D}_{12}(n) \\ \bm{D}_{21}(n) & \bm{D}_{22}(n) \end{bmatrix} \). In order to derive  \( \var\big( \wh{\Delta}_{\mathrm{regr}}(n) \big) \), we need to characterize the Gaussian vector \(\bm{z}(n) = \Big[ {\wh{\bm{\theta}}_S(n)}^\top \; {\wh{\bm{\theta}}_T}^\top \Big]^\top \).

    From~\autoref{eq:theta_T}, we have:
    \[
        \wh{\bm{\theta}}_T = \bm{A}_T^{-1}\bm{G}_T\bm{\theta}_T^\star+\bm{A}_T^{-1}\bm{Z}_T,
        \quad
        \wh{\bm{\theta}}_S(n) = \bm{A}_S(n)^{-1}\bm{G}_S(n)\bm{\theta}_S^\star+\bm{A}_S(n)^{-1}\bm{Z}_S(n),
    \]
    which gives us: \( \Big[ \wh{\bm{\theta}}_S(n)^\top \; \wh{\bm{\theta}}_T^\top \Big]^\top \sim \mathcal{N}\left( \bm{\mu}(n), \bm{\Sigma}(n) \right) \) with
    \[
        \bm{\mu}(n) = \Big[ \big( \bm{A}_S(n)^{-1}\bm{G}_S(n)\bm{\theta}_S^\star \big)^\top \; \big( \bm{A}_T^{-1}\bm{G}_T\bm{\theta}_T^\star \big)^\top \Big]^\top
        \;,\quad
        \bm{\Sigma}(n) =
        \begin{bmatrix}
            \sigma_S^2\,\bm{A}_S(n)^{-1}\bm{G}_S(n)\bm{A}_S(n)^{-1} & \bm{0}                                         \\
            \bm{0}                                                  & \sigma_T^2\,\bm{A}_T^{-1}\bm{G}_T\bm{A}_T^{-1}
        \end{bmatrix}.
    \]

    We conclude the computation of \( \var\big( \wh{\Delta}_{\mathrm{regr}}(n) \big) \) and the proof from the fact that:
    \[
        \var\Big[\wh{\Delta}_{\mathrm{regr}}(n) \Big] = \var\big( \bm{z}(n)^\top \; \bm{D}(n) \; \bm{z}(n) \big) = 2 \; \tr\Big( \big(\bm{D}(n) \Sigma(n) \big)^2 \Big) \;+\; 4 \; \bm{\mu}(n)^\top \bm{D}(n) \Sigma(n) \bm{D}(n) \; \bm{\mu}(n) \;.
    \]

\end{proof}

We prove the characterization of the classification transfer gain estimator:

\propb*

\begin{proof}[Proof of {\upshape\autoref{prop:expect_var_delta_hat_clf}}]
    Let
    \(
    \bm{Z}
    :=\Phi(-\bm{y}_T^{\mathrm{val}}\odot \wh{\bm{r}}_T)-\Phi(-\bm{y}_T^{\mathrm{val}}\odot \wh{\bm{r}}_c)\in\mathbb{R}^{n_T^{\mathrm{val}}},
    \quad\text{so that}\quad
    \wh{\Delta}_{\mathrm{clf}}(n)=\frac{1}{n_T^{\mathrm{val}}}\bm{1}^\top\bm{Z}.
    \)

    Condition on \(\wh{\bm{r}}_T,\wh{\bm{r}}_c\).
    For each \(i\), since \(y_{T,i}^{\mathrm{val}}\in\{\pm1\}\) with
    \(\mathbb{P}(y_{T,i}^{\mathrm{val}}=+1\mid \bm{X}_T^{\mathrm{val}})=\pi_i\),
    \begin{align*}
        \mathbb{E}[Z_i\mid \wh{\bm{r}}_T,\wh{\bm{r}}_c]
         & =\pi_i\Big(\Phi(-\wh r_{T,i})-\Phi(-\wh r_{c,i})\Big)
        +(1-\pi_i)\Big(\Phi(\wh r_{T,i})-\Phi(\wh r_{c,i})\Big)  \\
         & =\pi_i\,\wh d_{+,i}(n)+(1-\pi_i)\,\wh d_{-,i}(n)
        =:\wh m_i(n),
    \end{align*}
    hence \(\mathbb{E}[\bm{Z}\mid \wh{\bm{r}}_T,\wh{\bm{r}}_c]=\wh{\bm{m}}(n)\). Taking expectations and using linearity, \( \mathbb{E}\!\left[\wh{\Delta}_{\mathrm{clf}}(n)\right] =\frac{1}{n_T^{\mathrm{val}}}\bm{1}^\top \mathbb{E}[\bm{Z}] =\frac{1}{n_T^{\mathrm{val}}}\bm{1}^\top \mathbb{E}\!\left[\wh{\bm{m}}(n)\right]\) .

    By the law of total variance applied to the scalar \(\wh{\Delta}_{\mathrm{clf}}(n)\),
    \begin{align*}
        \var\!\left[\wh{\Delta}_{\mathrm{clf}}(n)\right]
         & =\mathbb{E}\!\left[\var\!\left(\wh{\Delta}_{\mathrm{clf}}(n)\,\middle|\,\wh{\bm{r}}_T,\wh{\bm{r}}_c\right)\right]
        +\var\!\left(\mathbb{E}\!\left[\wh{\Delta}_{\mathrm{clf}}(n)\,\middle|\,\wh{\bm{r}}_T,\wh{\bm{r}}_c\right]\right)    \\
         & =\frac{1}{(n_T^{\mathrm{val}})^2}\,
        \mathbb{E}\!\left[\bm{1}^\top \cov\!\big(\bm{Z}\mid \wh{\bm{r}}_T,\wh{\bm{r}}_c\big)\bm{1}\right]
        +\frac{1}{(n_T^{\mathrm{val}})^2}\,
        \var\!\left(\bm{1}^\top \wh{\bm{m}}(n)\right).
    \end{align*}
    The second term is
    \(
    \var(\bm{1}^\top \wh{\bm{m}}(n))=\bm{1}^\top \cov(\wh{\bm{m}}(n))\bm{1}.
    \)
    For the first term, by definition of \(\wh{\bm{v}}_{\mathrm{val}}(n)\), \( \wh{\bm{v}}_{\mathrm{val}}(n) =\diag\!\Big(\cov\!\big(\bm{Z}\mid \wh{\bm{r}}_T,\wh{\bm{r}}_c\big)\Big) \),
    hence \(\bm{1}^\top \cov(\bm{Z}\mid \wh{\bm{r}}_T,\wh{\bm{r}}_c)\bm{1}\) is the sum of its diagonal contribution
    \(\bm{1}^\top \wh{\bm{v}}_{\mathrm{val}}(n)\) and its off-diagonal contribution.
    Collecting terms yields the claimed characterization:
    \[
        \var\!\left[\wh{\Delta}_{\mathrm{clf}}(n)\right]
        =\frac{1}{(n_T^{\mathrm{val}})^2}\Big(
        \mathbb{E}\!\left[\bm{1}^\top \wh{\bm{v}}_{\mathrm{val}}(n)\right]
        +\bm{1}^\top \cov\!\big(\wh{\bm{m}}(n)\big)\bm{1}\Big).
    \]

    Moreover, for each \(i\), since \(y_{T,i}^{\mathrm{val}}\in\{\pm1\}\) with weights \(\pi_i\) and \(1-\pi_i\),
    \[
        \var(Z_i\mid \wh{\bm{r}}_T,\wh{\bm{r}}_c)
        =\pi_i\,\wh d_{+,i}(n)^2+(1-\pi_i)\,\wh d_{-,i}(n)^2-\wh m_i(n)^2,
    \]

\end{proof}

\section{Appendix: Lower bound of the transfer gain estimator}
\label{sec:supp_lower_bound}

We derive the lower-bound using the Bienaymé-Tchebychev inequality on the random variable $\wh{\Delta}_{\mathrm{regr}}(n)$, of expectation $\Delta_{\mathrm{regr}}^\star(n) + b(n, \lambda_c,\lambda_S,\lambda_T)$ and variance $\var \big( \wh{\Delta}_{\mathrm{regr}}(n) \big)$, we first restate~\autoref{prop:lower_bounds_transfer_gain}:

\propb*

\begin{proof}[Proof of {\upshape\autoref{prop:lower_bounds_transfer_gain}}]
    Let us consider the random variable $\wh{\Delta}_{\mathrm{regr}}(n)$ with expectation $\Delta_{\mathrm{regr}}^\star(n) + b(n, \lambda_c,\lambda_S,\lambda_T)$ and variance $\var(\wh{\Delta}_{\mathrm{regr}}(n))$.
    Cantelli's version of the Bienaymé–Chebyshev inequality states that for any $t>0$,
    \[
        \mathbb{P}\Big[ \wh{\Delta}_{\mathrm{regr}}(n) - \mathbb{E}\big[ \wh{\Delta}_{\mathrm{regr}}(n) \big]  \le -t \Big] \;\le\; \frac{\var \big( \wh{\Delta}_{\mathrm{regr}}(n) \big)}{\var \big( \wh{\Delta}_{\mathrm{regr}}(n) \big) + t^2} \quad\Longrightarrow\quad \mathbb{P}\Big [\wh{\Delta}_{\mathrm{regr}}(n) \ge \mathbb{E}\big[ \wh{\Delta}_{\mathrm{regr}}(n) \big] - t \Big] \;\ge\; \frac{t^2}{\var \big( \wh{\Delta}_{\mathrm{regr}}(n) \big) + t^2}.
    \]
    Choosing $t$ so that $\frac{t^2}{\var \big( \wh{\Delta}_{\mathrm{regr}}(n) \big) + t^2}=1-\delta$ gives $t^2=\var \big( \wh{\Delta}_{\mathrm{regr}}(n) \big) \; \frac{1-\delta}{\delta}$ and hence, with probability at least $1-\delta$,
    \[
        \wh{\Delta}_{\mathrm{regr}}(n) - \sqrt{\var\big( \wh{\Delta}_{\mathrm{regr}}(n) \big) \; \frac{1-\delta}{\delta}} - b(n, \lambda_c,\lambda_S,\lambda_T) \;\le\; \Delta_{\mathrm{regr}}^\star(n).
    \]

    We proceed similarly for the classification case.

\end{proof}

\section{Appendix: Theoretical analysis}
\label{sec:supp_theory}

We detail the proof of~\autoref{rm:beneficial-collab-large-n-ridge}, obtained by taking the expectation of \( \mathbb{E} \Big[ \Delta^\star(n) \Big] \) under the normalized Gaussian design assumption for the features.

\rma*

\begin{proof}[Proof of {\upshape\autoref{rm:beneficial-collab-large-n-ridge}}]

    On an independent validation design $X_T^{\mathrm{val}}$ with \iid $\mathcal{N}(\bm{0},\I_d)$ rows, the prediction error vectors are
    \begin{equation}\label{eq:err_vectors}
        e_T := \bm{X}_T^{\mathrm{val}}(\wh{\bm{\theta}}_T - \bm{\theta}_T^\star) \;,\qquad
        e_c := \bm{X}_T^{\mathrm{val}}(\wh{\bm{\theta}}_c - \bm{\theta}_T^\star) \;.
    \end{equation}
    With a fixed target and source training samples yields:
    \begin{align*}
        \mathbb{E}\big[\|e_T\|_2^2 \mid \bm{X}_T \big]           & = n_T^{\mathrm{val}} \Big( \lambda_T^2 \| \bm{A}_T^{-1} \bm{\theta}_T^\star \|_2^2 \;+\; \sigma_T^2\,\tr \big( \bm{A}_T^{-1} \bm{G}_T \bm{A}_T^{-1} \big)\Big) \;,                                                                                  \\
        \mathbb{E}\big[\|e_c\|_2^2 \mid \bm{X}_T, \bm{X}_S \big] & = n_T^{\mathrm{val}} \Big( \big\| \bm{A}_c(n)^{-1} \big( \bm{G}_S \bm{\Delta\theta}^\star - \lambda_c \bm{\theta}_T^\star\big) \big\|_2^2 \;+\; \tr\big( \bm{A}_c(n)^{-1}(\sigma_S^2 \bm{G}_S + \sigma_T^2 \bm{G}_T) \bm{A}_c(n)^{-1}\big)\Big) \;.
    \end{align*}

    We will now take the expectation toward the target and source training samples \(\bm{X}_T, \bm{X}_S\) with \iid \(\mathcal{N}(\bm{0},\I_d)\) rows. Invoking Marchenko-Pastur deterministic equivalents for the isotropic Gaussian design in the well-conditioned regime (\eg \(n_T \;, n \gg d\)):
    \[
        \bm{G}_T \;\approx\; n_T \I_d \;, \qquad \bm{G}_S \;\approx\; n \I_d \;, \qquad \bm{A}_T^{-1} \;\approx\; \frac{1}{n_T + \lambda_T} \I_d \;, \qquad \bm{A}_c(n)^{-1} \;\approx\; \frac{1}{n_T + n + \lambda_c} \I_d \;.
    \]
    Substituting these and combined~\autoref{eq:err_vectors} yields:
    \begin{align*}
        \mathbb{E}\Big[ \Delta^\star(n) \Big]
         & \approx n_T^{\mathrm{val}} \Big[ \lambda_T^2\,\frac{\| \bm{\theta}_T^\star\|_2^2}{(n_T + \lambda_T)^2}
            \;+\; \sigma_T^2 \; \frac{\tr(n_T \I_d)}{(n_T + \lambda_T)^2}
            \;-\; \frac{\|n \; \bm{\Delta\theta}^\star - \lambda_c \bm{\theta}_T^\star \|_2^2}{(n_T + n + \lambda_c)^2}
        \;-\; \frac{\tr \big((\sigma_S^2 n + \sigma_T^2 n_T) \I_d \big)}{(n_T + n + \lambda_c)^2} \Big]           \\
         & =       n_T^{\mathrm{val}} \lambda_T^2 \; \frac{\| \bm{\theta}_T^\star\|_2^2}{(n_T + \lambda_T)^2}
        \;+\; n_{\mathrm{val}} \sigma_T^2 \; \frac{d \; n_T}{(n_T + \lambda_T)^2}
        \;-\; n_T^{\mathrm{val}} \frac{\|n \; \bm{\Delta\theta}^\star - \lambda_c \bm{\theta}_T^\star\|_2^2}{(n_T + n + \lambda_c)^2}
        \;-\; n_T^{\mathrm{val}} \frac{d \; (\sigma_S^2 n + \sigma_T^2 n_T)}{(n_T + n + \lambda_c)^2} \;,
    \end{align*}

\end{proof}

\section{Appendix: Algorithms}
\label{sec:supp_alg}

\subsection{Appendix: Complete algorithm description}

We complete the description with a formal presentation in \autoref{alg:main_alg}.
\begin{algorithm}[H]
    \caption{Dynamic Positive Transfer}
    \label{alg:main_alg}
    {\small
        \begin{algorithmic}[1]
    \REQUIRE
    Sources $\{\{(\bm{x}_{s,i},y_{s,i})\}_{i=1}^{n_s}\}_{s=1}^{S}$; \
    target stats $(\bm{G}_T,\bm{A}_T^{-1},\wh{\bm{\theta}}_T)$; \
    noise $(\sigma_T,\sigma_S)$; parameter $\alpha$; \
    chunk size $c$; rounds $n_{\max}$; \
    init $\forall s$: $\bm{G}_S^{(s)}\!\gets\!\bm{0}$, $\bm{b}_S^{(s)}\!\gets\!\bm{0}$, $\bm{A}_S^{(s)}{}^{-1}$; \
    combined inverse $\bm{A}_c^{-1}$.

    \FOR{$n=1$ \textbf{to} $n_{\max}$}

    \STATE $\mathcal{S}_{\mathrm{act}} \gets \{\,s\in[S]: \text{source $s$ has samples}\,\}$
    \STATE $(\hat{s},\hat{\Delta}) \gets (\emptyset,-\infty)$

    \FOR{\textbf{each} $s \in \mathcal{S}_{\mathrm{act}}$}

    \STATE $\mathcal{I}_{s,n}\gets$ $c$ samples of source $s$
    \STATE $(\wt{\bm{G}},\wt{\bm{b}},\wt{\bm{A}}^{-1},\wt{\bm{A}}_c^{-1})\gets(\bm{G}_S^{(s)},\bm{b}_S^{(s)},\bm{A}_S^{(s)}{}^{-1},\bm{A}_c^{-1})$

    \FOR{\textbf{each} $(\bm{x},y)\in(\bm{X}_s[\mathcal{I}_{s,n}],\bm{y}_s[\mathcal{I}_{s,n}])$}

    \STATE $\wt{\bm{G}}\!\gets\!\wt{\bm{G}}+\bm{x}\bm{x}^\top$;\;
    $\wt{\bm{b}}\!\gets\!\wt{\bm{b}}+\bm{x}y$
    \STATE $\bm{k}\!\gets\!\wt{\bm{A}}^{-1}\bm{x}$;\ $\kappa\!\gets\!1+\bm{x}^\top\bm{k}$
    \STATE $\wt{\bm{A}}^{-1}\!\gets\!\wt{\bm{A}}^{-1}-\dfrac{\bm{k}\bm{k}^\top}{\kappa}$
    \STATE $\bm{k}_c\!\gets\!\wt{\bm{A}}_c^{-1}\bm{x}$;\ $\kappa_c\!\gets\!1+\bm{x}^\top\bm{k}_c$
    \STATE $\wt{\bm{A}}_c^{-1}\!\gets\!\wt{\bm{A}}_c^{-1}-\dfrac{\bm{k}_c\bm{k}_c^\top}{\kappa_c}$

    \STATE $\wt{\bm{\theta}}_S \gets \wt{\bm{A}}^{-1}\wt{\bm{b}}$;\quad
    $\Delta_s \gets \textsc{TrGain}\big(\alpha,..,\wt{\bm{\theta}}_S\big)$
    \STATE \textbf{if} $\Delta_s>\hat{\Delta}$ \textbf{then} $(\hat{s},\hat{\Delta})\gets(s,\Delta_s)$ and store $(\wt{\bm{G}},\wt{\bm{b}},\wt{\bm{A}}^{-1},\wt{\bm{A}}_c^{-1},\mathcal{I}_{s,n})$

    \ENDFOR
    \ENDFOR

    \STATE \textbf{Append} $\mathcal{I}_{\hat{s},n}$ to selected set
    \STATE $(\bm{G}_S^{(\hat{s})},\bm{b}_S^{(\hat{s})},\bm{A}_S^{(\hat{s})}{}^{-1},\bm{A}_c^{-1})
        \gets (\wt{\bm{G}},\wt{\bm{b}},\wt{\bm{A}}^{-1},\wt{\bm{A}}_c^{-1})$

    \ENDFOR
\end{algorithmic}

    }
\end{algorithm}

\subsection{Appendix: Complexity}

We first form the target Gram matrix and invert the regularized target Gram, which costs $\mathcal{O}(d^3+n_Td^2)$.
At each round, evaluating a candidate chunk of size $c$ requires $c$ rank-one updates, hence $\mathcal{O}(c\,d^2)$.
With the greedy strategy, this evaluation is performed for each active source, yielding $\mathcal{O}(|\mathcal{S}_{\mathrm{act}}|\,c\,d^2)$ per round in the worst case, while the uniform strategy evaluates a single source per round.
Across at most $n_{\max}$ admitted samples, the additional cost over target-only ridge is therefore on the order of $\mathcal{O}(n_{\max}d^2)$ for updates, and up to an extra factor proportional to the number of active sources for greedy scoring.

\section{Appendix: Experiments}
\label{sec:supp_exp}

\subsection{Appendix: Experimental setting of~\autoref{FIG:SAMPLE_SHARING_INTRO} and~\autoref{FIG:DELTA_ILLUS_EPS_0_2}}
\label{subsec:fig_intro_setting}

We consider the same setting than the synthetic benchmark~\autoref{subsec:simulation} with $\varepsilon=0.5$ for~\autoref{FIG:SAMPLE_SHARING_INTRO}.

\subsection{Appendix: Real data sets details}
\label{subsec:real_datasets_details}

We describe the datasets used in our experiments.

\paragraph{Email / Spambase (UCI):}  $9400$ emails with $9$ content and header features (word and character frequencies, capitalization patterns); the target is spam \vs ham.

\paragraph{Boston Housing:}  $4960$ instances with $4$ neighborhood/environmental variables (\eg RM, LSTAT, NOX); the target is the median value of the home.

We divide the data set into two subsets by clustering the samples, mimicking a realistic source–target partition.

\subsection{Appendix: Baseline details}
\label{subsec:baselines_details}

\paragraph{\citet{obst2021transfer}:} Transfer is implemented by fine-tuning the source estimator on the target loss: starting from the source weights, take \(k\) gradient steps with step size \(\alpha\) on the target squared-error objective.
Equivalently, the estimator applies a spectral filter \((\I_d-\alpha \bm{\Lambda})^{k}\) in the eigenbasis of the target covariance.
Hyperparameters \((\alpha,k)\) control the transfer strength and are selected in validation; if no gain is detected, they revert to the target-only model.

\paragraph{\citet{ChenOwenShi2014}:} They form a closed-form linear mixture of the source and target least-squares through a matrix weight \( \bm{W}(\lambda)=\bm{\Gamma}(\lambda)^{-1}\bm{\Psi}(\lambda) \) that depends on Gram matrices \(\bm{G}_T,\bm{G}_S\) and an auxiliary validation design covariance \(\bm{G}_T^{\mathrm{val}}\).
The single parameter \(\lambda>0\) controls regularization/transfer and is chosen by validation to minimize the validation error; the approach relies on the covariance structure (random-design) and the matrix inverses (e.g. \(\bm{G}_S^{-1}\)).

\paragraph{\citet{zhang2026unified}:} jointly optimizes source-sample weights and transferred quantities using an asymptotic KL-divergence generalization-error analysis; it typically uses all source data but reweights sources via convex optimization.

\paragraph{Hierarchical baseline:} a Bayesian hierarchical ridge where target and source coefficients share a latent mean $\mu$; estimate $\mu$ from both tasks, then compute the target MAP $\theta_t$ shrunk toward $\mu$.

\subsection{Appendix: Additional regression results on the main benchmarks}
\label{subsec:additional_results}

In this subsection, we gather additional results.

\paragraph{Effect of target sample size $n_T$} Adding the case $\varepsilon\in\{0,1\}$ (\autoref{fig:4_performance_comparison__eps_0}) yields the same pattern as \autoref{fig:synth_data_perf_exp}.
The magnitude of the gain increases, widening the margin over the target-only baseline.
Overall, this experiment replicates the earlier behavior and shows that our approach features greater transfer gains.

\begin{figure}[ht]
  \centering
  \begin{subfigure}{.48\columnwidth}
    \centering
    \includegraphics[width=0.45\linewidth]{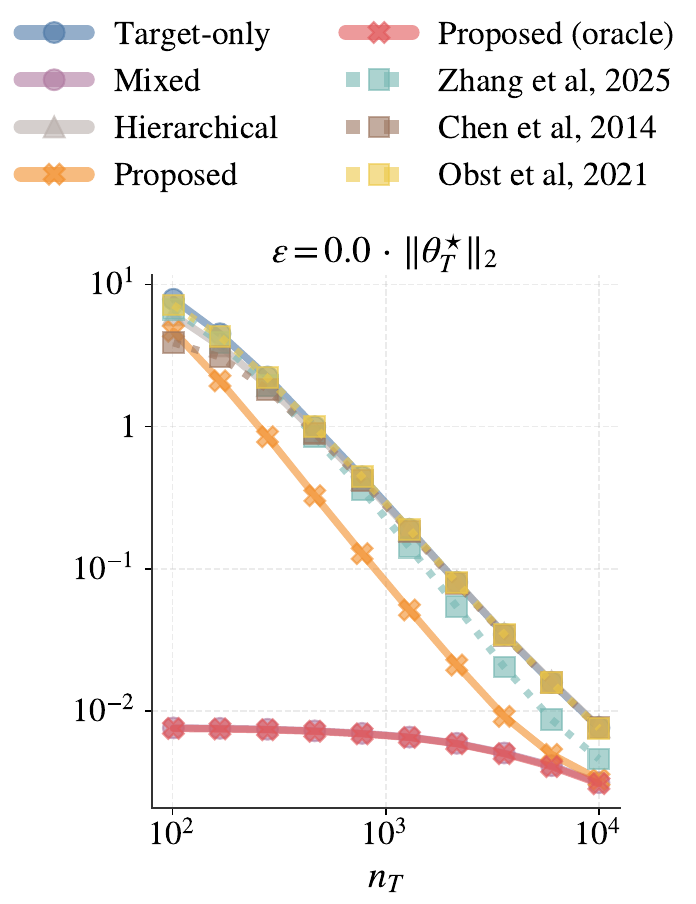}
    \subcaption{$\varepsilon = 0$}
  \end{subfigure}\hfill
  \begin{subfigure}{.48\columnwidth}
    \centering
    \includegraphics[width=0.45\linewidth]{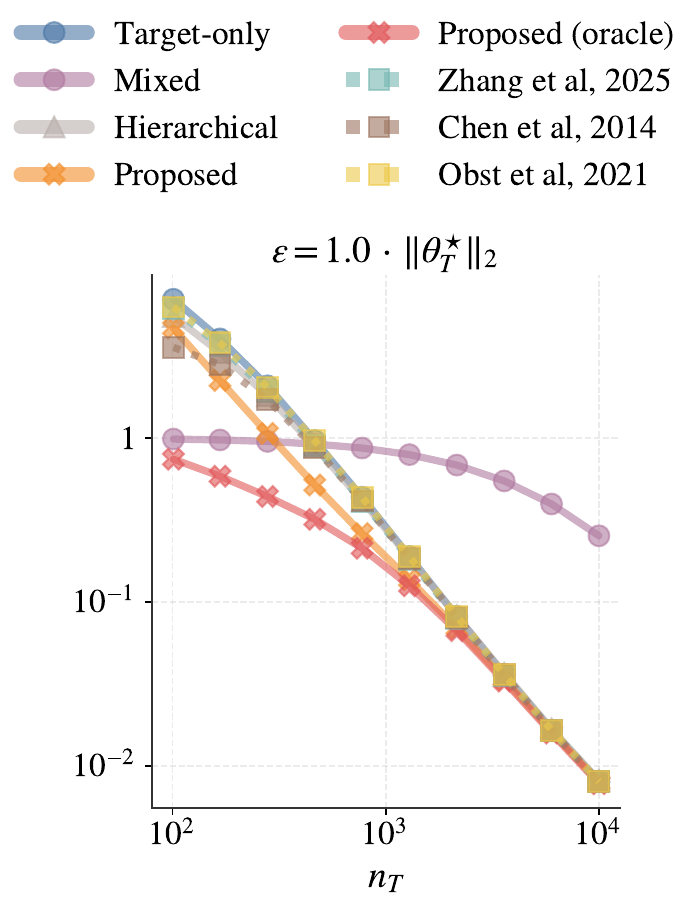}
    \subcaption{$\varepsilon = 1$}
  \end{subfigure}
  \caption{Predictive error comparison \wrt the number of target samples. The solid line reports the average; while the standard deviation is encoded in transparency.}
  \label{fig:4_performance_comparison__eps_0}
\end{figure}

We report the test-MSE additionally:

\begin{figure}[ht]
  \centering
  \begin{subfigure}{.24\columnwidth}
    \centering
    \includegraphics[width=0.95\linewidth]{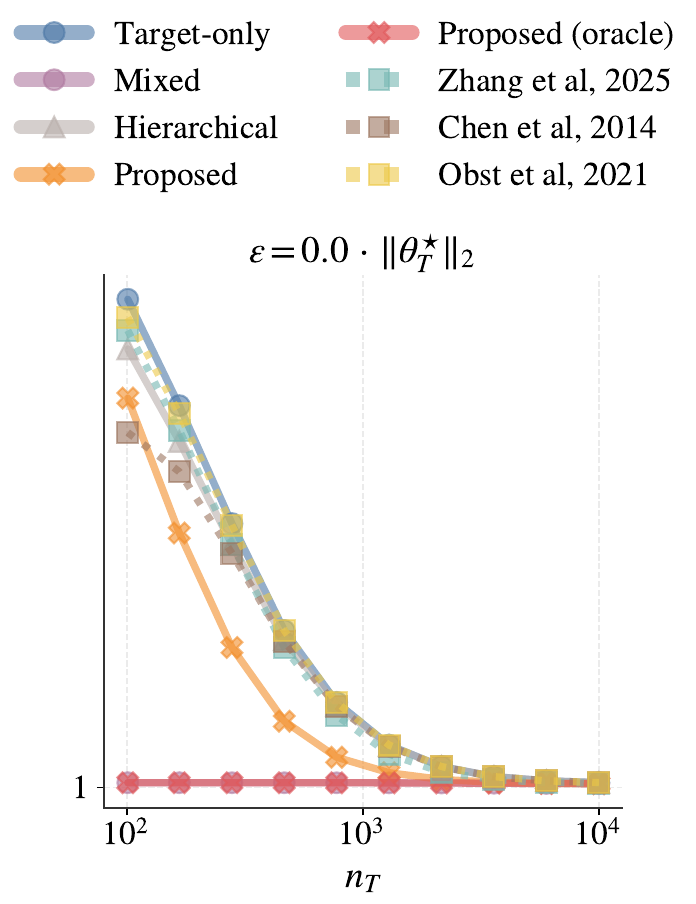}
    \subcaption{$\varepsilon = 0$}
  \end{subfigure}\hfill
  \begin{subfigure}{.24\columnwidth}
    \centering
    \includegraphics[width=0.95\linewidth]{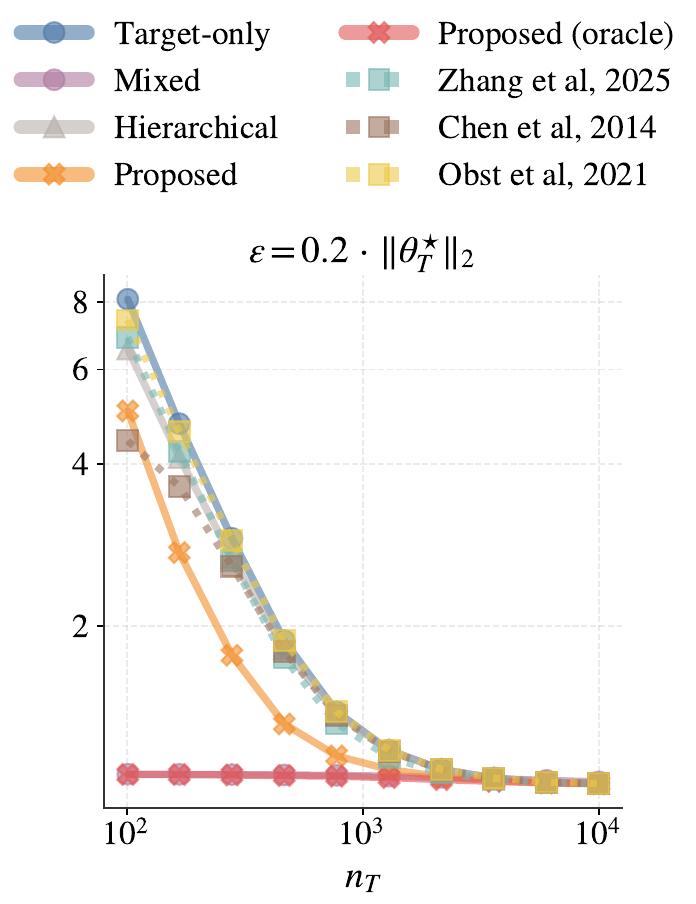}
    \subcaption{$\varepsilon = 0.2$}
  \end{subfigure}\hfill
  \begin{subfigure}{.24\columnwidth}
    \centering
    \includegraphics[width=0.95\linewidth]{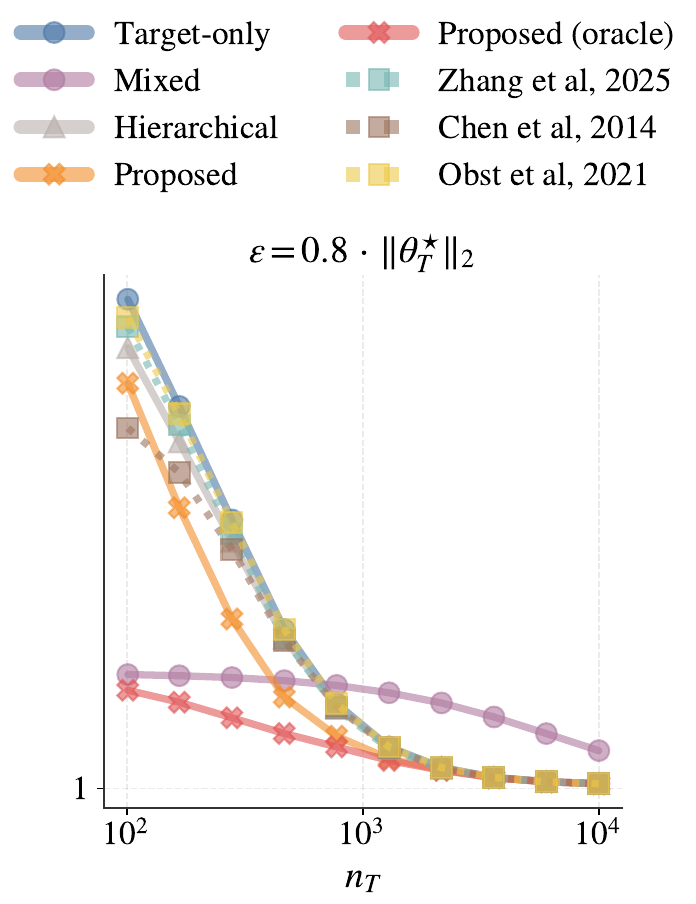}
    \subcaption{$\varepsilon = 0.8$}
  \end{subfigure}\hfill
  \begin{subfigure}{.24\columnwidth}
    \centering
    \includegraphics[width=0.95\linewidth]{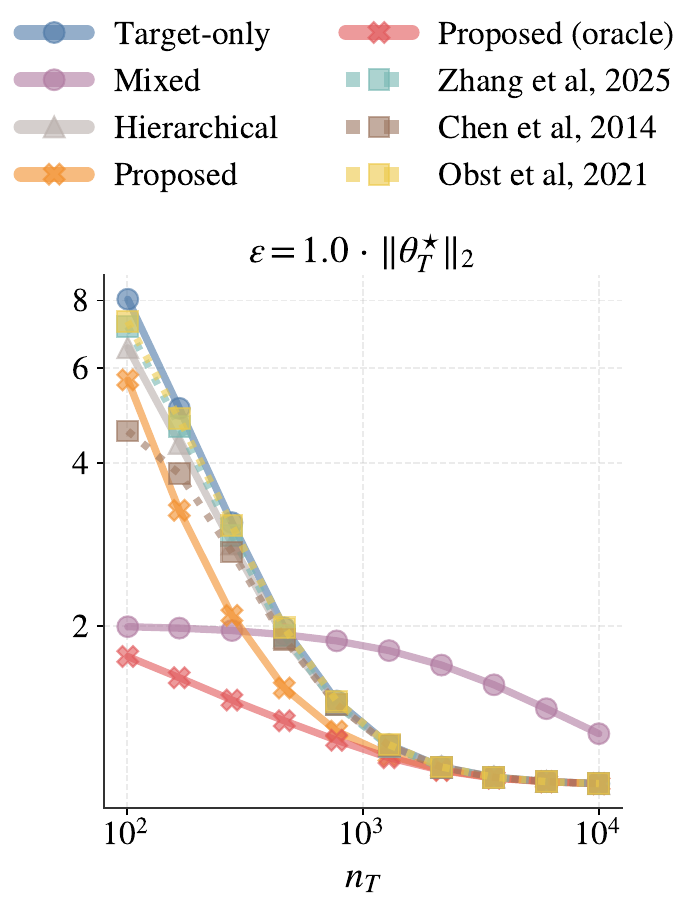}
    \subcaption{$\varepsilon = 1$}
  \end{subfigure}
  \caption{Predictive test-MSE comparison \wrt the number of target samples. The solid line reports the average; while the standard deviation is encoded in transparency.}
  \label{fig:4_performance_comparison__eps_0}
\end{figure}

\begin{figure}[ht]
  \centering
  \begin{subfigure}{.3\columnwidth}
    \centering
    \includegraphics[width=0.7\linewidth]{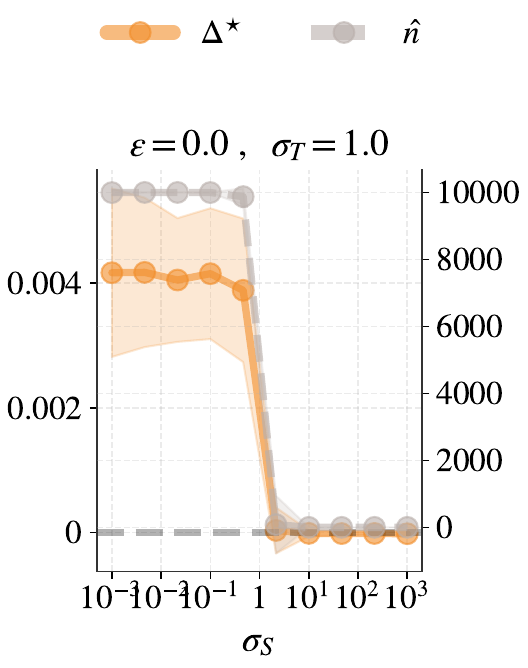}
    \subcaption{$\varepsilon = 0$}
  \end{subfigure}\hfill
  \begin{subfigure}{.3\columnwidth}
    \centering
    \includegraphics[width=0.7\linewidth]{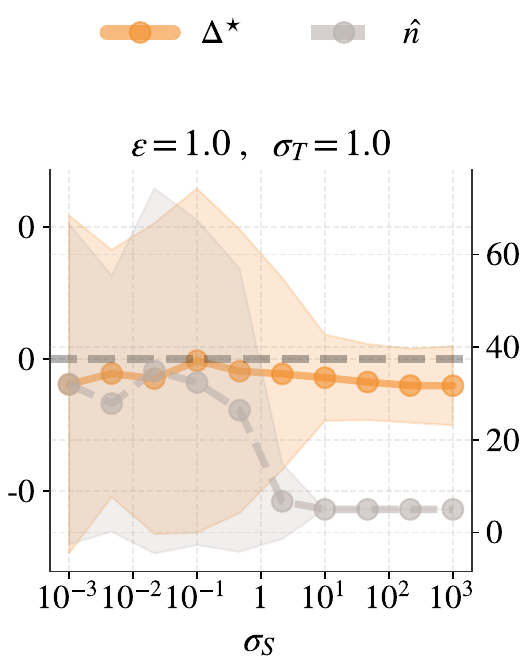}
    \subcaption{$\varepsilon = 1$}
  \end{subfigure}
  \begin{subfigure}{.39\columnwidth}
    \centering
    \caption{Predictive error comparison (left axis) and the number of samples borrowed \( \hat{n} \) (right axis) \wrt the source observation noise variance. The solid line reports the average; while the standard deviation is encoded in transparency.}
    \label{fig:02_sigma_influence__eps_0}
  \end{subfigure}
\end{figure}

\paragraph{Effect of target sample size $\sigma_S$:}  We also report $\varepsilon\in\{0,1\}$ in \autoref{fig:02_sigma_influence__eps_0}.
The shape of the overall curve mirrors \autoref{fig:02_sigma_influence__eps_0.2.pdf}; the larger $\varepsilon$ tends to shift the average empirical transfer gain until the sharing ends.

\subsection[Appendix: Effect of collaborative ridge parameter lambdac]{Appendix: Effect of collaborative ridge parameter $\lambda_c$}
\label{subsec:lambda_influence}

Moreover, we examine the effect of the collaborative ridge parameter.
We fix the number of target samples and the number of available source samples to the default along with the target and source noise variance set to \(\sigma_T=\sigma_S=1\).
We vary $\lambda_S$ from $0.01$ to $100.0$ and set $\lambda_c = \lambda_S + \lambda_T$.
We report the error $\mathrm{err}(\cdot)$ on the left axis and the number of samples borrowed \( \hat{n} \) on the right axis.

\begin{SCfigure}[1.1][ht]
  \centering
  \includegraphics[width=0.2\linewidth]{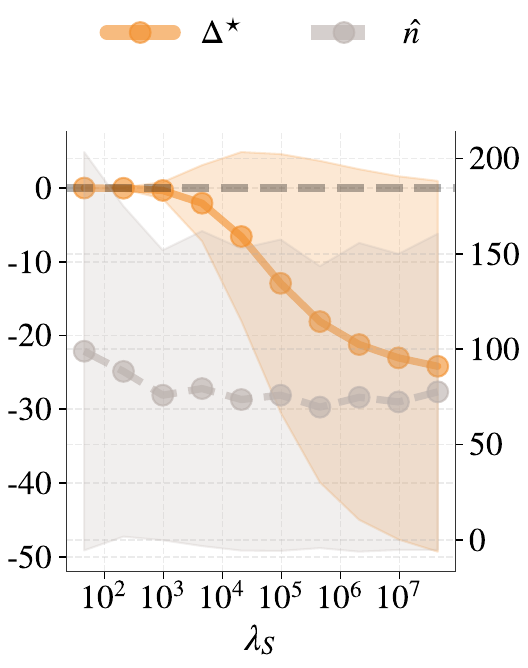}
  \caption{Predictive error comparison (left axis) and the number of samples borrowed \( \hat{n} \) (right axis) \wrt the collaborative ridge parameter $\lambda_S$. The solid line reports the average; while the standard deviation is encoded in transparency.}
  \label{fig:6_lbda_influence}
\end{SCfigure}

\autoref{fig:6_lbda_influence} plots the test error (left axis) and the selected number of borrowed source samples \( \hat{n} \) (right axis) as functions of the collaborative ridge \(\lambda_S\).
For large \(\lambda_S\), the collaborative estimator collapses toward \(\bm{0}\), \( \hat{n} \) drops, and the empirical transfer gain approaches a negative plateau.
In contrast, small \(\lambda_S\) enables positive transfer with a larger \( \hat{n} \).

\subsection{Appendix: Additional classification results on the main benchmarks}

In this subsection, we gather additional results.

\begin{figure}[ht]
  \centering
  \begin{subfigure}{.24\columnwidth}
    \centering
    \includegraphics[width=0.95\linewidth]{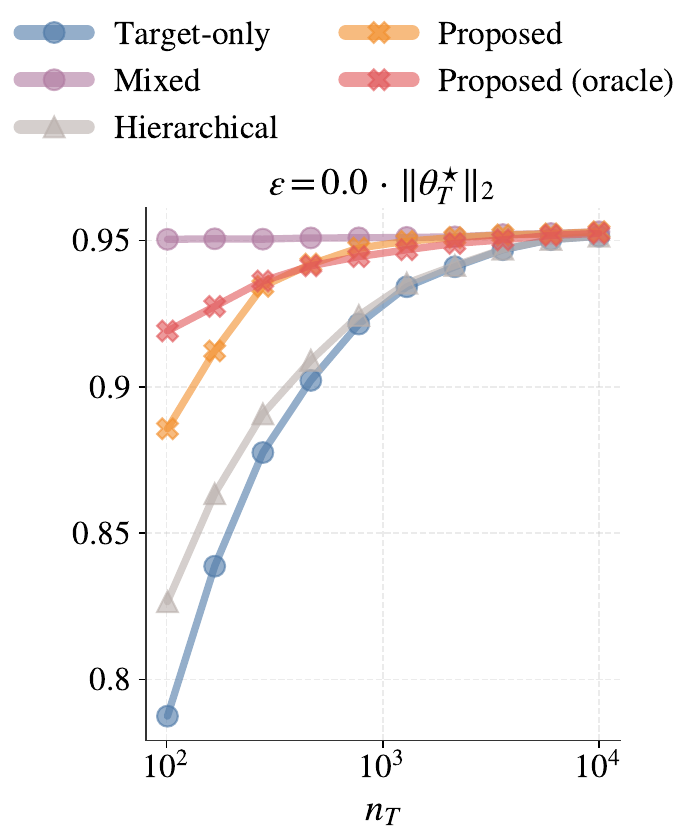}
    \subcaption{$\varepsilon = 0$}
  \end{subfigure}\hfill
  \begin{subfigure}{.24\columnwidth}
    \centering
    \includegraphics[width=0.95\linewidth]{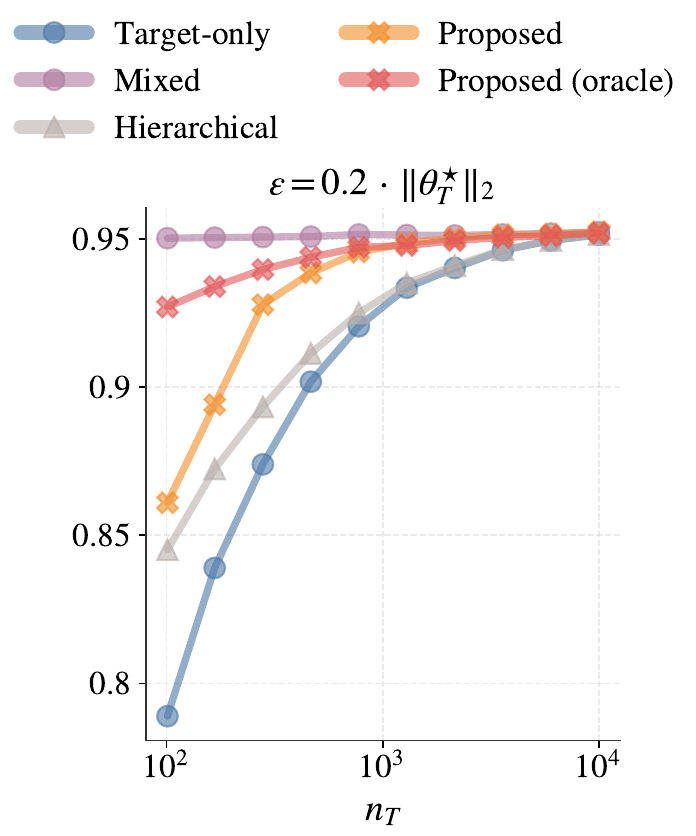}
    \subcaption{$\varepsilon = 0.2$}
  \end{subfigure}\hfill
  \begin{subfigure}{.24\columnwidth}
    \centering
    \includegraphics[width=0.95\linewidth]{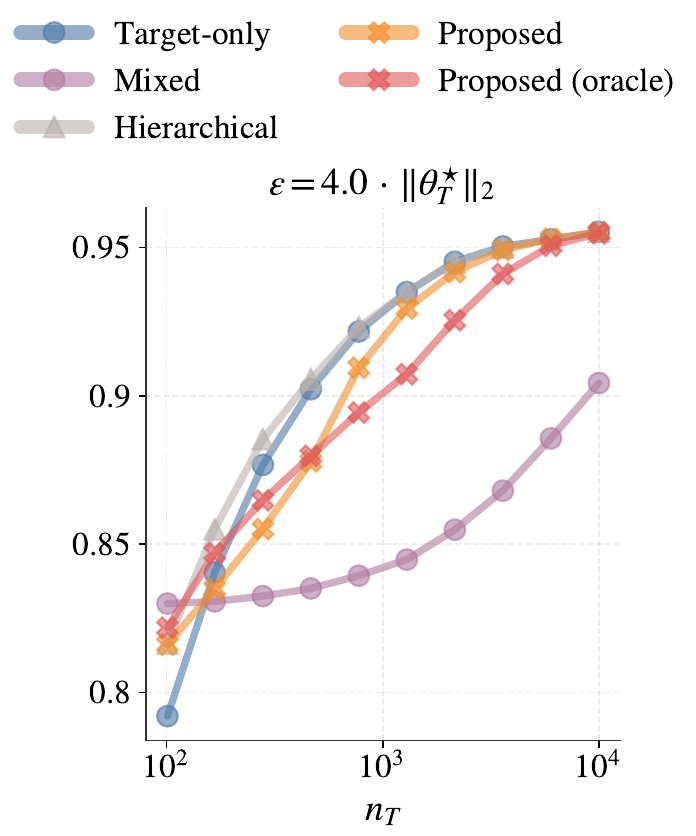}
    \subcaption{$\varepsilon = 0.8$}
  \end{subfigure}\hfill
  \begin{subfigure}{.24\columnwidth}
    \centering
    \includegraphics[width=0.95\linewidth]{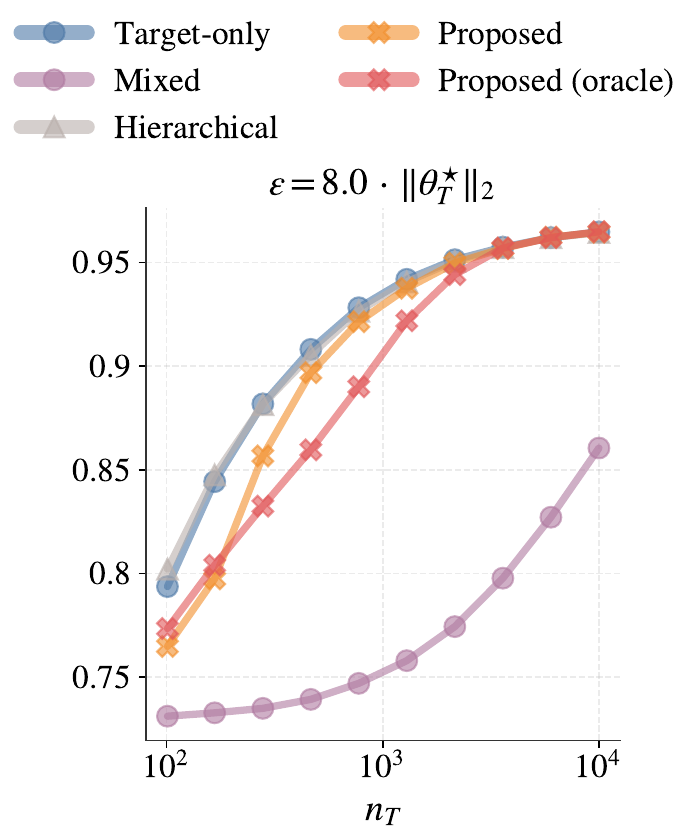}
    \subcaption{$\varepsilon = 1$}
  \end{subfigure}
  \caption{Predictive test-MSE comparison \wrt the number of target samples. The solid line reports the average; while the standard deviation is encoded in transparency.}
  \label{fig:09_performance_classif_acc__eps_8}
\end{figure}

\paragraph{Effect of target sample size $n_T$.} Including the classification setting (\autoref{fig:09_performance_classif_acc__eps_8}) reveals a different pattern than in regression (\autoref{fig:synth_data_perf_exp}).
When the model discrepancy is small, we recover behavior similar to the regression case; however, in this regime the \textsc{Mixed} baseline achieves the best performance.
In contrast, when the discrepancy is large enough for \textsc{Mixed} to degrade substantially, all other baselines perform comparably to the target-only model.
We conjecture that the classification setting effectively exhibits higher uncertainty than regression due to label quantization.
Heuristically, this is analogous to the regression case under very large noise, where transfer becomes more challenging.
Hence, in our experiments, the regression setting appears more informative in practice, as it better highlights the regimes where adaptive sample sharing can yield consistent gains.

\end{document}